\documentclass[review]{elsarticle}

\usepackage{times}
\usepackage{epsfig}
\usepackage{graphicx}
\usepackage{amsmath}
\usepackage{amssymb}
\usepackage{multirow}
\usepackage{makecell}
\usepackage{array}
\usepackage{pifont}
\usepackage{textcomp,booktabs}
\usepackage[usenames,dvipsnames]{color}
\definecolor{myred}{rgb}{.99,.91,.95}
\usepackage{color}
\usepackage{colortbl}
\newcommand{\tabincell}[2]{\begin{tabular}{@{}#1@{}}#2\end{tabular}}
\usepackage{lineno,hyperref}

\usepackage{tikz}
\newcommand*{\circled}[1]{\lower.7ex\hbox{\tikz\draw (0pt, 0pt)%
    circle (.5em) node {\makebox[1em][c]{\small #1}};}}

\newcommand{\xff}[1]{{\color{black} #1}}

\modulolinenumbers[5]

\journal{Pattern Recognition}

\begin{document}

\begin{frontmatter}

\title{
Boundary-induced and Scene-aggregated Network\\
for Monocular Depth Prediction}

\author[mymainaddress]{Feng Xue\fnref{eq}}
\author[mymainaddress]{Junfeng Cao\fnref{eq}}
\author[mysecondaryaddress]{Yu Zhou\corref{mycorrespondingauthor}}
\ead{yuzhou@hust.edu.cn}
\author[mymainaddress]{Fei Sheng}
\author[mymainaddress]{Yankai Wang}
\author[mymainaddress]{Anlong Ming}

\fntext[eq]{Equal contribution.}
\address[mymainaddress]{Beijing University of Posts and Telecommunications, Beijing, China}
\address[mysecondaryaddress]{Huazhong University of Science and Technology, Wuhan, China}
\cortext[mycorrespondingauthor]{Corresponding author}

\begin{abstract}
Monocular depth prediction is an important task in scene understanding.
It aims to predict the dense depth of a single RGB image.
With the development of deep learning,
the performance of this task has made great improvements.
However, two issues remain unresolved:
(1) The deep feature encodes the wrong farthest region in a scene,
which leads to a distorted 3D structure of the predicted depth;
(2) The low-level features are insufficient utilized,
which makes it even harder to estimate the depth near the edge with sudden depth change.
To tackle these two issues,
we propose the Boundary-induced and Scene-aggregated network (BS-Net).
In this network,
the Depth Correlation Encoder (DCE) is first designed to obtain the contextual correlations between the regions in an image,
and perceive the farthest region by considering the correlations.
Meanwhile, the Bottom-Up Boundary Fusion (BUBF) module is designed to extract accurate boundary that indicates depth change.
Finally, the Stripe Refinement module (SRM) is designed to refine the dense depth induced by the boundary cue,
which improves the boundary accuracy of the predicted depth.
Several experimental results on the NYUD v2 dataset and \xff{the iBims-1 dataset} illustrate the state-of-the-art performance of the proposed approach.
And the SUN-RGBD dataset is employed to evaluate the generalization of our method.
Code is available at https://github.com/XuefengBUPT/BS-Net.
\end{abstract}

\end{frontmatter}


\section{Introduction}
Dense and accurate depth is widely used in many computer vision applications,
such as autonomous driving \cite{LAF,LiStereo} and robotics \cite{JMOD2,XieTowards,LIU2020107112,YANG2020107141}.
Generally, acquiring dense depth depends on specific sensors,
i.e., stereo cameras \cite{YIN2017278,FIELDING2001531} and time-of-flight cameras.
To lower requirements of the sensor,
predicting dense depth from a single image attracts the attention of many researchers.

\begin{figure}
	\centering
	\includegraphics[width=1\linewidth]{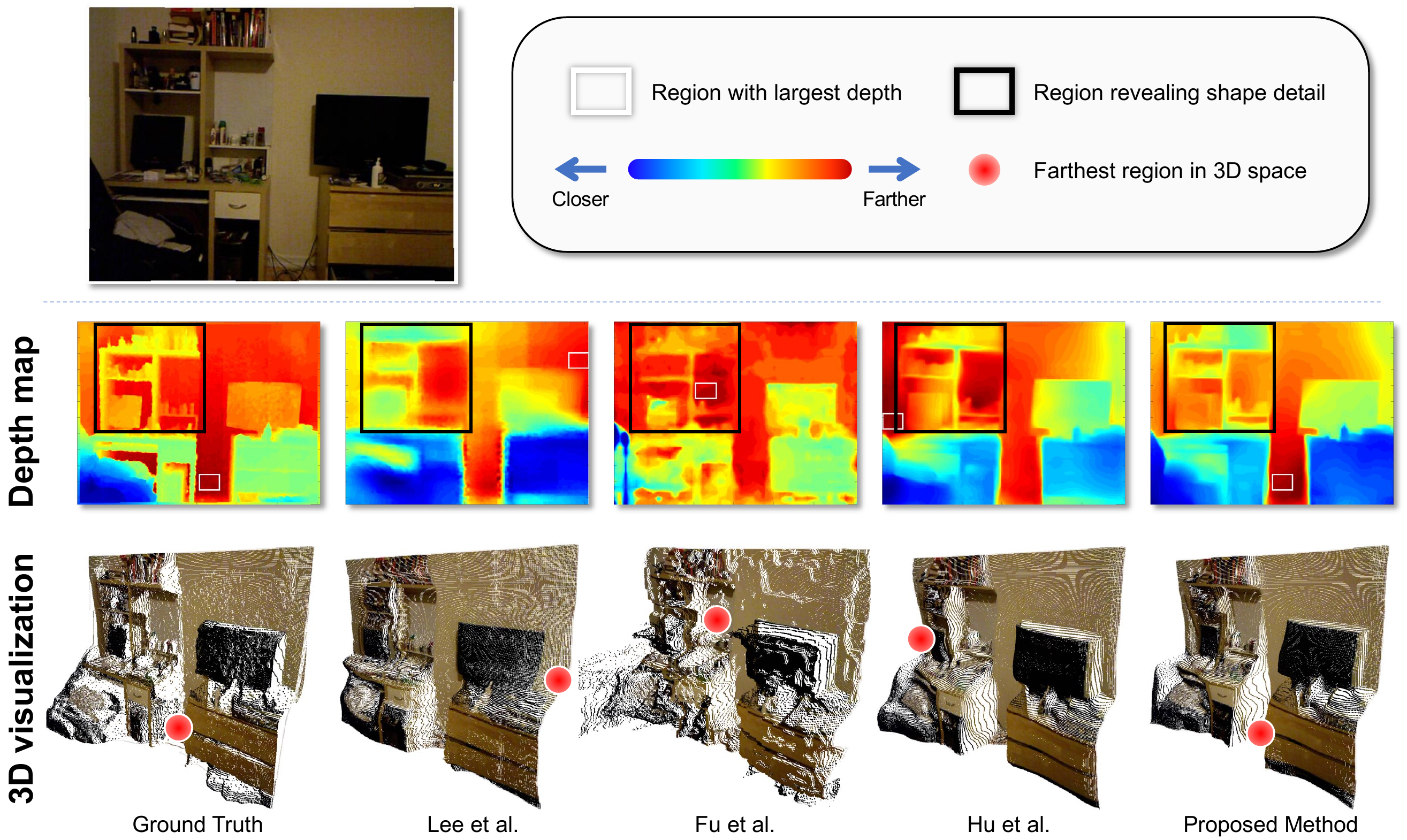}
		\caption{Visualization of the predicted depth.
		The white boxes indicate the farthest region in each depth map.
		The black boxes indicate the shape detail of the bookcase.}
	\label{fig:intro}
\end{figure}

In recent years,
many deep learning based methods are proposed to tackle this task
and achieve great performance gain.
Some of them \cite{Roy_Forest,Chen2019SARPN,revisit,Li_two_stream,Lee_2019_CVPR} are based on the encoder-decoder structure
and utilize specific modules or extra labels to improve the ability of their networks for scene understanding.
Although these methods achieve good performance in reducing pixel-wise depth error,
the ability to recover 3D space is still limited by the unstable 3D scene structure,
which is expressed as mistakenly estimating the farthest region of the scene.
As shown in Fig.\ref{fig:intro},
previous methods mistakenly regard the bookcase or the wall as the global farthest region,
leading to the 
scene distortion.
The intrinsic reason is that the encoder lacks consideration of the correlation between distant points when modeling the scene with high semantics,
thus each pixel in the high-level feature only represents the depth inside a region of the input image.
It results in that when the decoder is utilized to recover the pixel-wise depth,
the farthest region probably be wrongly estimated.
Besides,
the boundary of objects is an important cue in 3D space,
which indicates the sudden change of depth.
To obtain a depth map with sharp boundary,
several more recent methods \cite{revisit, Ramamonjisoa_2020_CVPR, Chen2019SARPN, Ramamonjisoa_2019_ICCV_Workshops} focus on improving the accuracy of the boundary in the predicted depth.
But due to the insufficient use of low-level features,
the boundaries are still hard to be recovered accurately.
The black polygon in Fig.\ref{fig:intro} depicts the detail of the bookcase.
Observably,
previous methods fail to recover the accurate shapes of the small objects in the bookcase.

In this paper,
the Boundary-induced and Scene-aggregated network (BS-Net) is proposed.
Firstly,
to perceive the farthest region,
we design the Depth Correlation Encoder (DCE).
On the one hand,
dilation convolutions are utilized to extract the correlation between long-distance pixels that are independent of each other, based on which the relative depth between different independent pixels is built.
On the other hand,
the Pyramid Scene Encoder (PSE) extracts the dominant features in multi-scale regions and fuses them to one,
which obtains the correlation between different regions.
Secondly,
to effectively recover the boundary,
we design the Bottom-Up Boundary Fusion (BUBF) module.
Starting from stage 2 of the encoder,
the module gradually fuses the feature of each two adjacent levels,
passing the location information to the high-level feature.
Thus, it eliminates the ineffective edges (indicating small depth change) in the low-level cues by the guidance of high-level cues,
and eventually extracts the boundary edges (indicating sudden depth change).
To effectively exploit the boundary,
the Stripe Refinement Module (SRM) is designed to replace the conventional regression module.
The convincing experiments on the NYUD v2 dataset \cite{nyu} demonstrate the effectiveness of our method,
and the additional results on the SUN-RGBD dataset \cite{sun} prove the generalization of our method.

The main contributions of our method can be summarized as follows:
\begin{itemize}
\item
To perceive the farthest region in a depth map,
the well-designed Depth Correlation Encoder (DCE) extracts the correlations between long-distance pixels and the correlations between multi-scale regions.
\item
To effectively extract and utilize the boundary,
the Bottom-Up Boundary Fusion (BUBF) module is designed to gradually fuse features of adjacent levels.
Meanwhile,
the Stripe Refinement Module (SRM) is designed to refine depth around the boundary.
\item Numerous experiments on the NYUD v2 dataset \cite{nyu} prove the effectiveness of our method.
The proposed method achieves state-of-the-art performance.
And the experiment on the SUN-RGBD dataset \cite{sun} proves the generalization of our method.
\end{itemize}

\section{Related Work}
\subsection{Monocular Depth Prediction}
\noindent
{\bf{Handcraft-based depth prediction methods}} 
utilize the handcrafted features to estimate the depth from a single image.
Saxena \emph{et al.} \cite{saxena2005learningdepth} design a multi-scale Markov Random Field to predict the depth 
and extend it with multiple cues combination \cite{4531745}. 
Liu \emph{et al.} \cite{Liu2014} define the task as a discrete-continuous optimization via Conditional Random Field.
However,
handcrafted features limit the expression of depth and are unable to predict accurate depth.

\noindent
{\bf{Deep learning based depth prediction methods}} significantly improve the performance of depth prediction.
Initially,
the encoder-decoder structure is widely utilized 
and the deep learning based depth prediction is divided into two parts \cite{Eigen,Fergus}:
the global coarse depth prediction and the local shape detail refinement.
The former obtains a low-resolution depth map as an intermediate product,
while the latter obtains a final depth map with original-resolution.
To obtain smooth and accurate depth,
Hu \emph{et al.} \cite{revisit} utilize the multi-level features and multi-task loss.
With a similar goal,
Chen \emph{et al.} \cite{Chen2019SARPN} gradually refine the final depth.
Hao \emph{et al.} \cite{Hao_Detail} reserve the high resolution of feature maps by utilizing continuous dilated convolutions.
Zhang \emph{et al.} \cite{Zhang2018Deep} retain the object shape by hierarchically fusing the side-output with the decoder.
Dhamo \emph{et al.} \cite{Dhamo2018Peeking} propose the Layered Depth
Image to perceive the occluded scene regions.
Some others focus on optimizing the absolute depth by first predicting the relative depth between pixels or regions.
Fu \emph{et al.} \cite{DORN} introduce the spacing-increasing discretization to regress the ordinal.
Similarly,
Lee \emph{et al.} \cite{Lee_2019_CVPR} refine the final map by combining the predicted relative depth of various scales.
Besides,
multi-task learning is widely concerned.
Several methods \cite{PADHY2019165,Fergus,jiao,Peng2015,Li2018Monocular} jointly utilize the labels of depth, semantic segmentation, and surface normal in predicting monocular depth.
However, existing encoder-decoder based methods always focus on how to obtain detailed shapes of objects 
and ignore the false prediction of the farthest region,
which leads to the distortion of the 3D structure.
Our method focuses on addressing this problem.

\noindent
{\bf{The datasets for depth prediction}}
are divided into two categories according to the scene,
namely indoor datasets and outdoor datasets.
In the datasets of indoor scenes \cite{nyu,sun,scannet},
only the NYUD v2 dataset \cite{nyu} provides dense and accurate depth and aligns with the corresponding RGB images.
Thus, it is utilized to evaluate the performance of monocular depth prediction methods.
Other datasets provide raw dense depth and RGB pairs captured by Kinect.
Thus, they are utilized to evaluate the generalization.
Since our method uses gradients to express the boundary in the depth map,
we utilize the NYUD v2 dataset to verify the performance of our method,
and other datasets to verify the generalization of our method.
For the datasets of outdoor scenes \cite{4531745,KITTI,cityscapes},
KITTI \cite{KITTI} and CityScapes \cite{cityscapes} provide the RGB images captured by the camera and the corresponding sparse depth captured by the 3D Lidar.
But since the sparse depth cannot be used to extract pixel-wise gradients,
these two datasets are not employed to verify our method.
Mono3D \cite{4531745} provides dense but inaccurate depth.
Thus, it is not employed either.

\subsection{Deep Feature}
\noindent 
{\bf{Contextual Extractor}} is significant for scene understanding tasks,
e.g., monocular depth prediction, semantic segmentation, and object detection.
Chen \emph{et al.} \cite{Chen} employ the Atrous Spatial Pyramid Pooling (ASPP) to encode the image context at multi-scale.
Fu \emph{et al.} \cite{DORN} propose the Scene Understanding Modular to achieve a comprehensive understanding of the input image.
Liu \emph{et al.} \cite{Liu2015ParseNet} extract the global context information by global pooling and combine it with a simple Fully Convolutional Network structure.
Different from other methods,
to predict the correct farthest region in the depth map,
our method extracts two types of contextual features.

\noindent
{\bf{Multi-level Feature Fusion}} exploits the features from different levels of the encoder,
And it is widely utilized in many computer vision tasks \cite{8099508,8578510,revisit,Lu2019Occlusion,Hao_Detail,CHEN201990,DUTTA202064}.
Hu \emph{et al.} \cite{revisit} concatenate all side-outputs of the encoder to fuse multiple level features.
Xu \emph{et al.} \cite{8099508} learn the relationship between features of different levels
and extend it with an attention model for multi-level feature fusion \cite{8578510}.
Hao \emph{et al.} \cite{Hao_Detail} gradually fuse the features of different levels.
Different from other methods,
our motivation is to reduce the gap of multi-level features in the fusion process.
Thus,
starting from the lowest level,
our method gradually fuses features of each two adjacent levels.

\noindent 
{\bf{Boundary Cue}} reveals the boundaries of objects.
Thus,
it has a high response at the depth discontinuity between foreground and background
and affects the accuracy of depth prediction around the object boundary.
Different from other work \cite{OLP,IS,Lu2019Occlusion,ROH2007931,ZHANG2020107484} which utilize the boundary as the label,
we lack the label of boundary.
Thus,
this paper applies the gradient of depth to guide the generation of boundary cue.

\section{Method}
\begin{figure}
	\centering
	\includegraphics[width=1\linewidth]{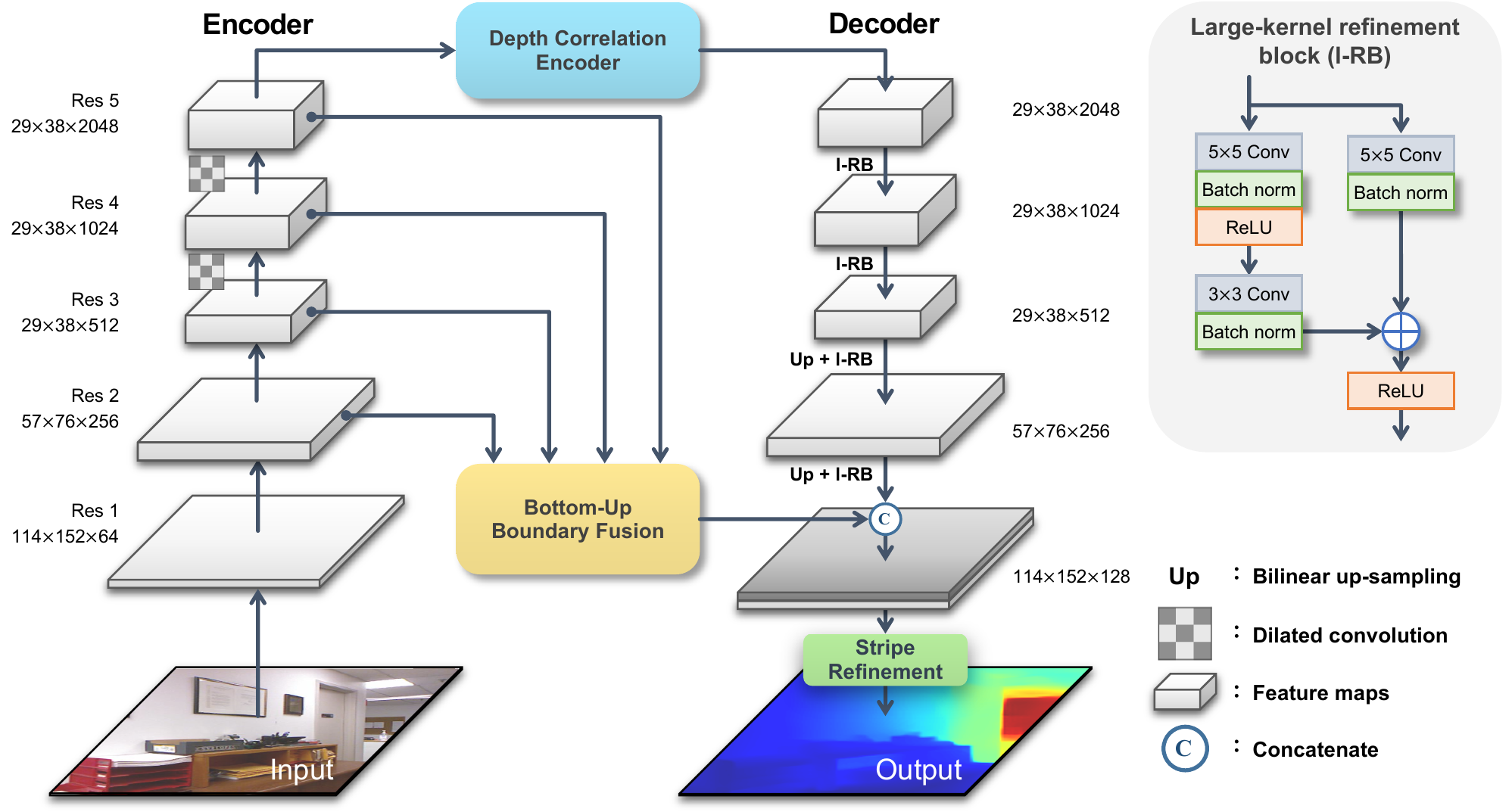}
	\caption{
		Our network structure,
		which consists of an encoder-decoder network,
		a Depth Correlation Encoder (DCE),
		a Bottom-Up Boundary Fusion (BUBF) module and a Stripe Refinement Module (SRM).}
	\label{fig:pipeline}
\end{figure}

In this section,
we introduce the Boundary-induced and Scene-aggregated Network (BS-Net).
This network consists of an encoder, a decoder, and three modules.
As shown in Fig.\ref{fig:pipeline},
the network firstly employs the ResNet \cite{7780459} as the encoder.
To obtain a larger receptive field,
the down-sampling operation of stage 4 and 5 are replaced by the dilated convolution \cite{dilated}.
Then,
to perceive the farthest region,
the Depth Correlation Encoder (DCE) is designed to aggregate the correlations between long-distance pixels and multi-scale regions from Res 5 (see Sec.\ref{sec:DCE}).
Subsequently,
a decoder is designed to predict the depth with the learned feature.
It consists of four consecutive steps.
The first two steps utilize the Large-Kernel Refinement Blocks (l-RB) to compress channels and keep the resolution.
The last two steps utilize the combined operation of upsampling and l-RB,
which is equal to the UpProjection in \cite{7785097},
to expand the resolution to $\frac{1}{2}$ of the original size.
Meanwhile, 
taking the side-output of Res 2/3/4/5 as input,
the Bottom-Up Boundary Fusion (BUBF) module is introduced to gradually extract boundaries indicating sudden changes in depth (see Sec.\ref{sec:BUBF}).
Finally,
the features from the decoder and BUBF are concatenated.
A Stripe Refinement Module (SRM) is designed to refine the high-resolution depth by the guidance of the learned boundary (see Sec.\ref{sec:SU}).

\subsection{Depth Correlation Encoder}
\label{sec:DCE}
\begin{figure}
	\centering
	\includegraphics[width=1\linewidth]{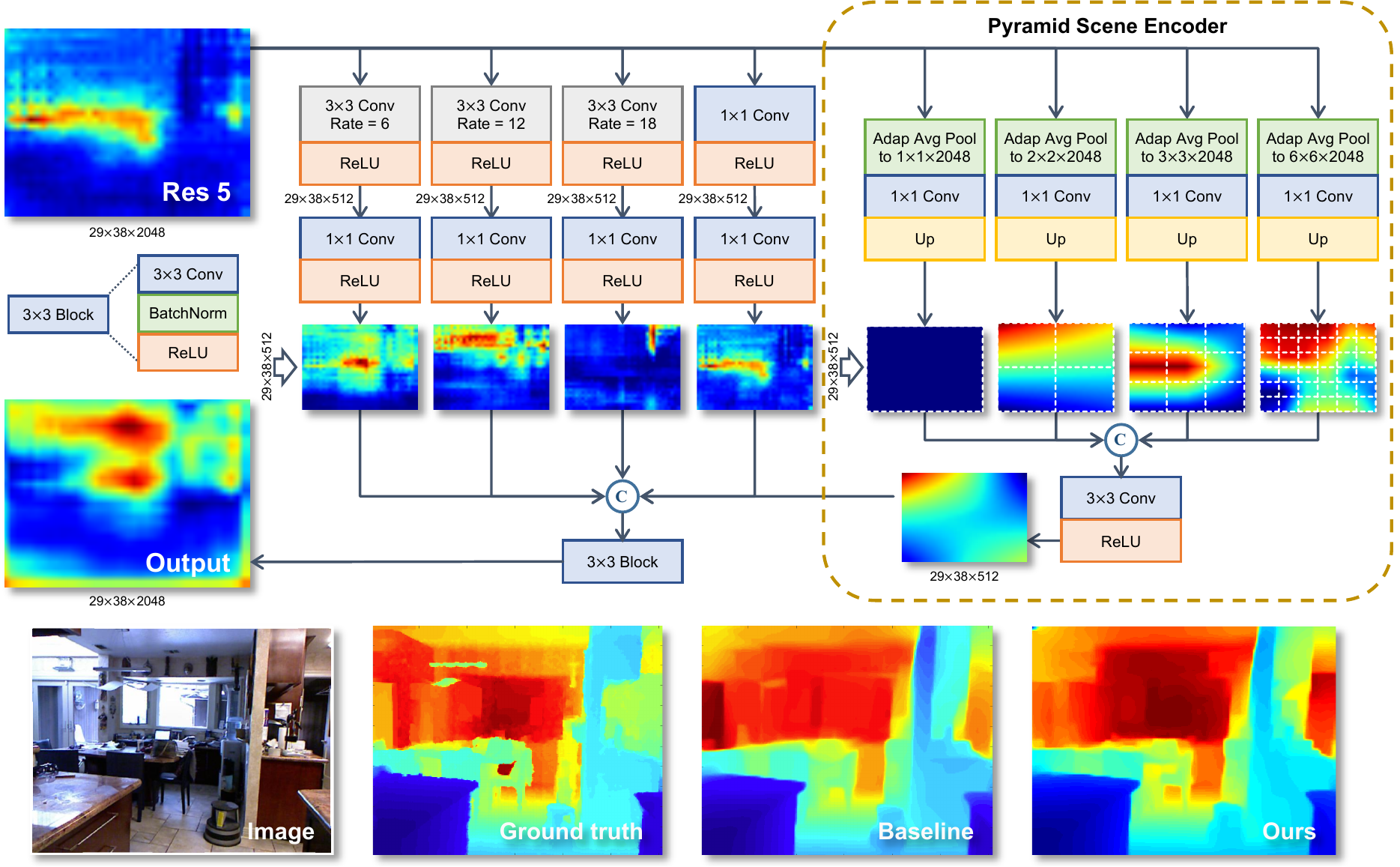}
	\caption{
	Illustration of our proposed Depth Correlation Encoder (DCE). The black arrow indicates that the feature maps have the same channel number.}
	\label{fig:DCE}
\end{figure}

To perceive the farthest region,
the Depth Correlation Encoder (DCE) is proposed to capture the correlations between long-distance pixels and multi-scale regions.
Given the features of Res 5 as input,
the module captures the correlations by eight parallel branches
and perceives the farthest region by considering these correlations,
as depicted in Fig. \ref{fig:DCE}.
Specifically,
the first three branches utilize three parallel dilated convolutions respectively.
They all have the same kernel size of $3\times3$,
but different dilated rates, i.e., 6, 12, 18.
For the contextual feature, namely Res 5 ,
each pixel encodes the depth in a region of the input image.
Thus, these dilated convolutions encode the correlation between two distant regions of the input image.
The largest-rate kernel has a receptive field of the entire image,
while the smallest one only covers about $\frac{1}{9}$ of the entire image.
The multi-scale correlations between pixels are extracted sufficiently.
After each dilated convolution,
a $1\times1$ convolution is employed to integrate the information between channels
and remove the grid artifacts caused by the dilated convolution.
As shown in Fig.\ref{fig:DCE},
the farthest region perceived by dilated convolutions is different from Res 5 
and is more consistent with the true depth.
The fourth branch is two consecutive $1\times1$ convolutions,
learning local cross-channel interaction.

The last four branches construct the Pyramid Scene Encoder (PSE) to encode the correlations between multi-scale regions and locate the farthest regions by considering the correlations.
In detail,
by utilizing the adaptive average pooling operation,
these four branches respectively downsample the Res 5 to four sizes ($n\times n \times 2048|n\in\{1,2,3,6\}$).
Let $H_{res5}$ and $W_{res5}$ denote the height and width of Res 5,
$s_H$ and $s_W$, namely the pooling strides in the vertical and horizontal directions,
are equal to $\lfloor\frac{H_{res5}}{n}\rfloor$ and $\lfloor\frac{W_{res5}}{n}\rfloor$,
meanwhile the kernel sizes of pooling in the two directions are equal to $H_{res5}-(n-1)\times s_{H/W}$ and $W_{res5}-(n-1)\times s_{H/W}$, respectively.
For clarity,
the four paths corresponding to $n=1/2/3/6$ are denoted as 1-st, 2-nd, 3-rd, and 4-th paths respectively.
Since each pixel of the pooled feature is the average of the Res 5 feature inside a region of size $\frac{H_{res5}}{n}*\frac{W_{res5}}{n}$,
abnormally large values are smoothed by other surrounding pixels.
Then,
for features of all paths,
a $1\times1$ convolution is used to combine the feature of each channel and reduce the channel number to $512$.
Subsequently,
the features of all paths are upsampled to a fixed size $29\times38\times512$ by UpProjection \cite{7785097} and concatenated.
The features of different scale regions encode the dominant features in their regions because of the average pooling operation.
Finally, a $3\times3$ convolution is employed to learn the correlations between the multi-scale regions 
and further reduce the channel number to $512$.
By considering the depth information of different scale regions and their correlations,
PSE extracts the relative depth change between regions.
As the visualizations of the last four paths shown in Fig.\ref{fig:DCE},
the farthest region is in the upper left bin,
and the feature in the 4-th path provides location information of the farthest region.

Eventually,
all the five branches of DCE are concatenated,
a final $3\times3$ convolution is employed to fuse the contextual correlations of depth, and the channel is reduced to $2048$.
The effectiveness of DCE is shown in Fig.\ref{fig:DCE}.
The response of the lower half of Res 5 is low,
which is consistent with the ground truth.
However,
the left of Res 5 has high response,
which leads to the wrong prediction of the farthest region (see the dark red left side of baseline in Fig.\ref{fig:DCE}) 
and is completely different from the true depth.
By encoding the correlation between the long-distance points and the feature inside regions,
it is observed that the farthest region is consistent with the true depth.

\subsection{Bottom-Up Boundary Fusion} 
\label{sec:BUBF}
\begin{figure}
	\centering
	\includegraphics[width=1\linewidth]{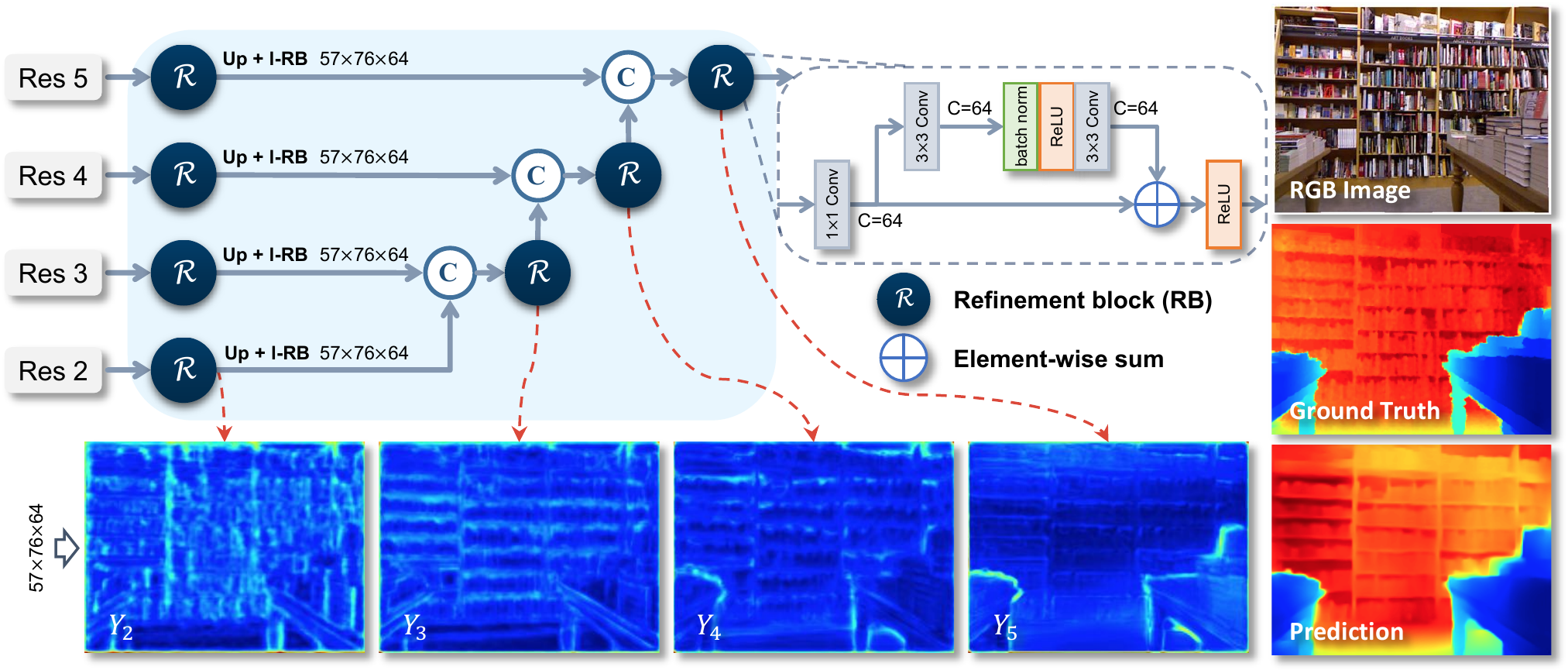}
	\caption{
		Illustration of the proposed BUBF module and the mean of channel maps ($Y_2,Y_3,Y_4,Y_5$) in four levels.
		The black arrow indicates that the feature maps have the same channel number.
		}
	\label{fig:BUBF}
\end{figure}

To accurately recover the depth around the boundary,
the Bottom-Up Boundary Fusion (BUBF) module is designed to extract the boundary (indicating sudden changes in depth) and remove the non-boundary pixel (indicating smooth changes in depth).
As shown in Fig.\ref{fig:BUBF},
this module is based on the complementarity of low-level and high-level features:
Low-level features are rich in location information of edge
but lack semantic information,
while high-level features are rich in semantic information.
The boundary of objects is encoded but inaccurately located.
Since the sudden change in depth is generally not inside the object area,
but near the boundary of objects,
the proposed module delivers the low-level location information to the higher level layer by layer,
fusing the feature of two adjacent levels 
to obtain the accurate boundary.

Specifically,
the whole module is composed of four steps.
In the first step,
all side-outputs, i.e., Res 2,3,4,5,
are refined by a well-designed Refinement Block (RB).
In more detail,
the RB first uses a $1\times1$ convolution to reduce the channel number to 64,
which integrates the information across all channels.
Then, the following residual block refines the feature.
In the second step,
to align the features of multi-levels,
the refined features in all levels are upsampled to the original resolution of the input image
and further improved by the l-RB.
This operation is the same as the UpProjection in \cite{7785097}.
In the third step,
beginning from the lowest level (corresponding to Res 2),
the cue of each level is concatenated with that of the adjacent higher level for latter fusion.
In the fourth step,
the cues of two adjacent levels are fused by utilizing another RB.
And in the same way,
the BUBF module iteratively combines the refined cues of two adjacent levels,
until the highest level (corresponding to Res 5) is reached.

Since the final loss function computes the gradient of depth (see Sec \ref{sec:LF}),
which indicates the depth boundary, 
the boundaries indicating depth changes are encoded by the extracted features.
Therefore,
the high-level features (such as Res 5) encode the boundary with depth changes,
but lack pixel-level location information.
In contrast,
the low-level features (such as Res 2) contain sufficient location information,
but fail to reveal the depth changes.
The BUBF passes the accurate location information to the high-level features from stage 2 to stage 5,
making the depth boundary to be located more accurately.
As shown in Fig.\ref{fig:BUBF},
compared to the initial feature at the lowest level (namely $Y_2$ in Fig.\ref{fig:BUBF}),
the feature combined by adjacent levels suppresses pixels with smooth depth change,
as shown in $Y_3,Y_4,Y_5$ in Fig.\ref{fig:BUBF}.
The function $R()$ denotes the RB,
the function $up()$ denotes the upsampling,
and the function $lR()$ denotes the l-RB.
$X_i$ is the side-output of Res $i$ .
$Y_i$ is the fused feature map of each level.
The first fused feature map $Y_2$ is stated as $Y_2=up(R(X_2))$
and other fused feature $Y_i$ is formulated as:
\begin{equation}
 Y_i=R\Big(lR\big(up(R(X_i))\big)\circled{C}Y_{i-1}\Big)\,, i\in\{3,4,5\}
\end{equation}
where $\circled{C}$ denotes the concatenation operation.
$Y_5$ is the output of the whole module.

\subsection{Depth Prediction}
\label{sec:SU}
The previous method \cite{revisit} exploits three $5 \times 5$ convolutions to predict the final depth map from the decoder.
However,
the small-kernel convolution causes two problems because of its limited receptive field.
\begin{itemize}
 \item It only aggregates the local feature at each pixel, making the local confusion in depth prediction inevitable.
 \item It fails to take full advantage of the boundary and the global contextual features.
\end{itemize}

\begin{figure}
	\centering
	\includegraphics[width=0.9\linewidth]{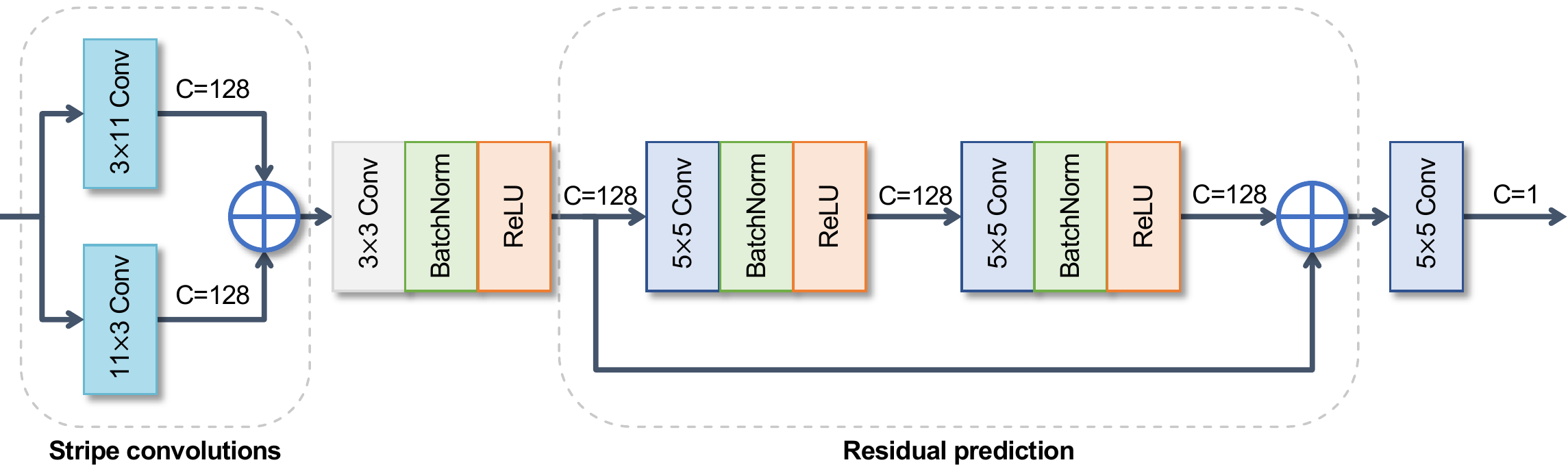}
	\caption{
		The structure of our Stripe Refinement Module.}
	\label{fig:IM}
\end{figure}

To address these issues,
a Stripe Refinement Module (SRM) is proposed to refine the depth across the boundary by utilizing the stripe convolution.
Specifically,
after the channel-wise concatenation of the decoder module and the BUBF,
the obtained features are taken as the input of the regression module which consists of two steps,
as illustrated in Fig.\ref{fig:IM}.
In the first step,
two stripe convolutions (with kernel sizes $3\times 11$ and $11\times3$) are utilized to aggregate the pixels nearby the boundary,
in a wide range of both vertical and horizontal directions.
Since the global contexts along orthogonal directions make significant contributions to indicating the relative depth,
the depth change between the object and its background is better recognized.
Secondly, a $3\times3$ convolution is employed to fuse the features extracted by two stripe convolutions.
Thirdly,
three convolutions with kernel size $5\times5$ are employed to refine the final depth map.
Especially,
to predict the depth map more accurately,
the fused feature before convolutions are delivered to the last convolution by a skip connection.

\textbf{Differences with whole strip masking (WSM) \cite{Heo_2018_ECCV} and vertical pooling \cite{Gan_2018_ECCV}:}
Intuitively,
the structure of SRM looks like the whole strip masking \cite{Heo_2018_ECCV} and vertical pooling \cite{Gan_2018_ECCV}.
However, SRM is substantially different to them.
Firstly,
we use a $3\times11$ convolution and a $11\times3$ convolution,
instead of $3\times W$ convolution and $H\times3$ convolution,
or $H\times1$ pooling operation,
where $W$ and $H$ denote the width and height of the input feature.
Secondly,
the element-wise sum is employed to fuse the features obtained by two strip convolutions.
Thirdly,
the strip convolution is employed to refine the depth near the boundary acquired by BUBF to improve the depth,
instead of exploiting the trend of the scene \cite{Heo_2018_ECCV},
or vertically aggregate image feature \cite{Gan_2018_ECCV}.

\subsection{Loss Function}
\label{sec:LF}
To train the network,
a ground truth depth map in the training data is denoted as $\textbf{G}$
and its corresponding prediction is denoted as $\textbf{P}$.
Each pixel in the ground truth depth map is denoted as $\textbf{g}_i\in\textbf{G}$, and $\textbf{p}_i\in\textbf{P}$ for the prediction.
Following \cite{revisit},
the loss function of our network is composed of three terms,
i.e.,
pixel-wise depth difference $l_{depth}$,
gradient difference $l_{grad}$,
and surface normal difference $l_{normal}$.
Assuming that $\nabla_x()$ and $\nabla_y()$ denote the spatial gradient of a pixel in $x$ and $y$ directions,
the surface normal of a ground truth depth map and its predicted depth map are denoted as $\textbf{n}_i^\textbf{p}=[-\nabla_x(\textbf{p}_i),-\nabla_y(\textbf{p}_i),1]$ and $\textbf{n}_i^\textbf{g}=[-\nabla_x(\textbf{g}_i),-\nabla_y(\textbf{g}_i),1]$.
The three loss functions can be formulated as follows:
\begin{itemize}
    \item $l_{depth} = \frac{1}{N_p}\sum^{N_p}_{i=1}\ln(\|\textbf{p}_i-\textbf{g}_i\|_1+\alpha)$
    \item $l_{gradient} = \frac{1}{N_p}\sum^{N_p}_{i=1}\ln\Big(\nabla_x(\|\textbf{p}_i-\textbf{g}_i\|_1+\alpha)+\nabla_y(\|\textbf{p}_i-\textbf{g}_i\|_1+\alpha)\Big)$
    \item $l_{normal} = \frac{1}{N_p}\sum^{N_p}_{i=1}\Big(1-\frac{	\left<\textbf{n}_i^\textbf{g},\textbf{n}_i^\textbf{p}\right>}{\sqrt{\left<\textbf{n}_i^\textbf{g},\textbf{n}_i^\textbf{g}\right>}\sqrt{\left<\textbf{n}_i^\textbf{p},\textbf{n}_i^\textbf{p}\right>}}\Big)$
\end{itemize}
where $\left<.,.\right>$ denotes the inner production, $N_p$ denotes the pixel number in an image, $\alpha$ is a hyper parameter, which is set to $0.5$.
For the overall loss of the BS-Net,
the weights of these three loss functions are equal:
\begin{equation}
 l_{overall} = l_{depth}+l_{normal}+l_{grad}
\end{equation}
where $l_{overall}$ is the overall loss function.
Since the pixels around the boundary have large gradients of depth,
the gradient difference $l_{grad}$ naturally guides the BUBF module to learn the boundary.

\section{Experiment}
In this section,
we evaluate the proposed method on three challenging datasets.
Firstly,
the NYUD v2 dataset \cite{nyu} is employed to evaluate the performance of depth and edge.
Secondly,
the iBims-1 dataset \cite{ibims} is employed to evaluate the quality of depth boundaries and other relevant metrics.
Thirdly,
the SUN-RGBD dataset \cite{sun} is employed to evaluate the generalization of our method.

\subsection{Implementation Detail}
Our proposed network is implemented using the NVIDIA 1080Ti GPUs and PyTorch framework.
The ResNet-50 is adopted as the backbone network
and initialized by the pre-trained model on ILSVRC \cite{ImageNet}.
The classification layers for output are removed.
To preserve the size of feature maps,
 dilated convolutions are employed in the last two stages of the backbone.
Other parameters in the decoder, DCE, BUBF, and SRM are randomly initialized.
We train our model for 20 epochs and set batch size to 8.
The Adam optimizer is adopted with parameters $(\beta{1},\beta{2})=(0.9,0.999)$.
The weight decay is $10^{-4}$.
The initial learning rate is set to 0.0001
and reduced by 10\% for every 5 epochs.

Following the previous work \cite{Chen2019SARPN,revisit},
we consider 654 RGB-D pairs for testing and 50k pairs for training.
The data augmentation is performed on the training images in the same way as \cite{revisit}.
To train the model,
all images and labels are downsampled to $320\times240$ pixels from the original size ($640\times480$) using bilinear interpolation,
and then cropped to $304\times228$ pixels from the central part.
To align with the network output,
the cropped labels are downsampled to $152\times114$ pixels.
Furthermore,
the output of the network is upsampled to $304\times228$ pixels in the testing process to evaluate the model.
Note that,
we do not clip the depth maps of predicted depth to a fixed range.

\subsection{Metrics}
Three kinds of metrics are employed to thoroughly evaluate the proposed method.

\noindent
\textbf{Depth accuracy:}
Let $N$ denotes the total number of pixels in the test set,
$\overline{d_i}$ denotes pixel $i$ on the predicted depth,
and $d_i^*$ denotes pixel $i$ on the true depth.
Following \cite{DORN,revisit,8578297},
four metrics indicating pixel-wise accuracy are employed:
(1) Mean absolute relative error (REL): ${\frac{1}{N}}{\sum_{i}{\frac{|{\bar{d}_{i}}-{d_{i}^{*}}|}{{d_{i}^{*}}}}}$.
(2) Mean log 10 error (log10): ${\frac{1}{N}}{\sum_{i}{\|{{{log_{10}}\bar{d}_i}-{{log_{10}}{d_i^*}}}}\|}$.
(3) Root mean squared error (RMS): $\sqrt{{\frac{1}{N}}{\sum_{i}{({\bar{d}_{i}}-{d_i^*})^{2}}}}$.
(4) Accuracy under threshold $t_d$: $max({\frac{d_i^*}{\bar{d}_i}},{\frac{\bar{d}_i}{d_i^*}})=\delta<t_d(t_d\in[1.25,1.25^2,1.25^3])$.

\noindent
\textbf{Boundary accuracy in predicted depth:}
Following \cite{revisit},
we measure the boundary accuracy in predicted depth.
The Sobel operator is used to recover boundary from the predicted $\textbf{P}$ and the true depth maps $\textbf{G}$,
obtaining $\textbf{P}_{sobel}$ and $\textbf{G}_{sobel}$
The pixels larger than threshold $t_e$ ($t_e\in\{0.25, 0.5, 1\}$) is boundary,
$tp$ is the number of correct boundary pixels.
$fp$ and $fn$ correspond to the false positive and false negative.
The precision $P=\frac{tp}{tp+fp}$, recall $R=\frac{tp}{tp+fn}$ and F1 score $\frac{2\times P\times R}{P+R}$ are used to evaluate the performance.


\noindent
\textbf{Depth boundary error and other relevant metrics:}
Several novel metrics are introduced in \cite{ibims},
which are used to evaluate our algorithm more comprehensively.
In more detail,
the depth boundary errors (DBEs) measures the completeness and accuracy of the boundary in the predicted dense depth map.
In addition, we employ two other depth error metrics in \cite{ibims}:
planarity error (PE) to measure the depth accuracy in 3D space,
directed depth errors (DDEs) to measure the proportions of too far and too close predicted depth pixels.

\noindent
\textbf{Normalized distance error of farthest region:}
To evaluate the farthest region in the predicted depth,
another metric is introduced.
The maps $\textbf{P}$ and $\textbf{G}$ are partitioned into $m\times m$ rectangular regions with the same size.
And the mean depth inside each region is calculated.
In map $\textbf{P}$,
the region with largest mean depth is located as $\textbf{P}_{max} = (u_1,v_1)$,
and $\textbf{G}_{max} = (u_2,v_2)$ corresponds to that of the map $\textbf{G}$,
where $u_i,v_i \in \mathbb{N^+}$ and $1 < u_i,v_i < m$.
The normalized distance error of the farthest region is stated as the distance between $\textbf{P}_{max}$ and $\textbf{G}_{max}$:
\begin{equation}
 E = \frac{1}{\mathcal{N}_{test}}\sum\nolimits^{\mathcal{N}_{test}}_{n=1}\frac{1}{m\sqrt{2}}\|\textbf{P}^n_{max}-\textbf{G}^n_{max}\|_2
\end{equation}
where $\mathcal{N}_{test}$ is the number of test images,
$n$ indicates the $n$-th depth map,
and $\frac{1}{m\sqrt{2}}$ is used to normalize the error distance.

\subsection{NYUD v2 dataset}
The NYUD v2 dataset \cite{nyu} includes 120K pairs of RGB and depth maps captured by a Microsoft Kinect,
and is split into a training set (249 scenes) and a test set (215 scenes).
The images in this dataset have a resolution of $640\times480$.
The labeled data contains 1449 RGB-Depth pairs which are split into two parts:
795 pairs for training and 654 pairs for testing.
We firstly conduct analysis and comparison with several state-of-the-art methods on this dataset quantitatively and qualitatively.
Secondly, we conduct an in-depth analysis of our method by designing sufficient ablation studies,
evaluating all metrics of evaluation for depth prediction and scene edges on depth maps.

\begin{table}[!tb]
\footnotesize
\begin{center}
\begin{tabular}{|c|c|ccc|ccc|}
\Xhline{1.2pt}
\multirow{2}{*}{Method}&\multirow{2}{*}{backbone}&\multicolumn{3}{c|}{higher is better ($\delta$\textless)}& \multicolumn{3}{c|}{lower is better}\\
&&$1.25$ & $1.25^2$ & $1.25^3$ & REL & RMS & log10 \\
\hline
\hline
Saxena \emph{et al.} \cite{4531745} &-& 0.447 & 0.745 & 0.897 & 0.349 & 1.214 & - \\
Karsch \emph{et al.} \cite{Karsch} & -& - & - & - & 0.35 & 1.20 & 0.131 \\
Liu \emph{et al.} \cite{Liu2014} & - & - & - & - & 0.335 & 1.06 & 0.127 \\
 Xu \emph{et al.} \cite{8099508} & ResNet-50 & 0.811 & 0.954 & 0.987 & 0.121 & 0.586 & 0.052 \\
Eigen \emph{et al.} \cite{Eigen} & - & 0.611 & 0.887 & 0.971 & 0.215 & 0.907 & - \\
Eigen \emph{et al.} \cite{Fergus} & VGG & 0.769 & 0.950 & 0.988& 0.158 & 0.641 & - \\
Dharmasiri \emph{et al.} \cite{8205954} & VGG & 0.776 & 0.953 & 0.989& 0.156 & 0.624 & - \\
 Laina \emph{et al.} \cite{7785097} & ResNet-50 & 0.811 & 0.953 & 0.988& 0.127 & 0.573 & 0.055 \\
Lee \emph{et al.} \cite{8578140} & ResNet-152 & 0.815 & 0.963 & 0.991& 0.139 & 0.572 & - \\
 Hu \emph{et al.} \cite{revisit} & ResNet-50 & 0.843 & 0.968 & 0.991& 0.126 & 0.555 & 0.054 \\
 Xu \emph{et al.} \cite{8578175} & ResNet-50 & 0.817 & 0.954 & 0.987&\textcolor{red}{0.120} & 0.582 & 0.055 \\
Fu \emph{et al.} \cite{DORN} & ResNet-101 & 0.828&0.965&0.992&0.115& 0.509&0.051 \\
Lee \emph{et al.} \cite{Lee_2019_CVPR} & DenseNet-162 & 0.837&0.971&0.994&-&0.538&- \\
Chen \emph{et al.}* \cite{Chen2019SARPN} & ResNet-50 &0.853&0.959&0.991&0.121&0.545&0.052 \\
 Ramamonjisoa \emph{et al.} \cite{Ramamonjisoa_2019_ICCV_Workshops} & ResNet-50 &\textcolor{red}{\textbf{0.888}}&\textcolor{red}{ \textbf{0.979}}&\textcolor{red}{\textbf{0.995}}&0.139&\textcolor{red}{0.495}&\textcolor{red}{\textbf{0.047}} \\
Yin \emph{et al.} \cite{0Enforcing} & ResNeXt-101 &0.875&0.976&0.994&\textbf{0.108}&\textbf{0.416}&0.048 \\
Swami \emph{et al.} \cite{icip} & SENet-154 &0.870&0.974&0.993&0.115&0.528&0.049 \\
\hline
 \textbf{Ours} & ResNet-50 & 0.846 & 0.969&0.992& 0.123 & 0.550 & 0.053 \\
\Xhline{1.2pt}
\end{tabular}
\end{center}
\caption{Depth accuracy and error of different methods on the NYUD v2 dataset.
The bold type indicates the best performance.
The red number indicates the best performance with the same backbone of ResNet-50.
* Using a ResNet-50 backbone,
which is different from the original paper.}
\label{table:depth}
\end{table}

\begin{table}[!htp]
\footnotesize
\begin{center}
\begin{tabular}{|c|c|c|ccc|}
\hline
Thres & Method & Backbone & Precision & Recall & F1-score \\
\hline
\hline
\multirow{12}{*}{\textgreater{}0.25} 
& Eigen \emph{et al.} \cite{Eigen} & - & 0.346 & 0.322 & 0.323\\
\specialrule{0em}{-1pt}{-1pt}
& Eigen \emph{et al.} \cite{Fergus} & VGG & 0.544 & 0.481 & 0.500\\
\specialrule{0em}{-1pt}{-1pt}
& Dharmasiri \emph{et al.} \cite{8205954} & VGG &0.577&\textbf{0.626}&\textbf{0.591}\\
\specialrule{0em}{-1pt}{-1pt}
& Laina \emph{et al.} \cite{7785097} & ResNet-50 & 0.489 & 0.435 & 0.454 \\
\specialrule{0em}{-1pt}{-1pt}
& Xu \emph{et al.} \cite{8099508} & ResNet-50 & 0.516 & 0.400 & 0.436 \\
\specialrule{0em}{-1pt}{-1pt}
& Fu \emph{et al.} \cite{DORN} & ResNet-101& 0.320 & 0.583 & 0.402 \\
\specialrule{0em}{-1pt}{-1pt}
& Hu \emph{et al.} \cite{revisit} & ResNet-50 & 0.635 & 0.480 & 0.540 \\
\specialrule{0em}{-1pt}{-1pt}
& Lee \emph{et al.} \cite{Lee_2019_CVPR} & DenseNet-162 & 0.475 & 0.354 & 0.390 \\
\specialrule{0em}{-1pt}{-1pt}
& Chen \emph{et al.}* \cite{Chen2019SARPN} & ResNet-50 & \textbf{\color{red}0.650} & \color{red}0.483 & {\color{red}0.548} \\
\specialrule{0em}{-1pt}{-1pt}
&  Ramamonjisoa \emph{et al.} \cite{Ramamonjisoa_2019_ICCV_Workshops} & ResNet-50 &0.416&0.353&0.374 \\
 \specialrule{0em}{-1pt}{-1pt}
 & Yin \emph{et al.} \cite{0Enforcing} & ResNeXt-101 &0.523&0.459&0.480\\
 \specialrule{0em}{-1pt}{-1pt}
 & Ours & ResNet-50 & 0.644 & {\color{red}0.483} & 0.546 \\
 \specialrule{0em}{-1pt}{-1pt}
 \hline
 \specialrule{0em}{-1pt}{-1pt}
\multirow{12}{*}{\textgreater{}0.5} 
 & Eigen \emph{et al.} \cite{Eigen} & - &0.443& 0.278 & 0.327\\
 \specialrule{0em}{-1pt}{-1pt}
 & Eigen \emph{et al.} \cite{Fergus} &VGG& 0.587 & 0.456 & 0.501\\
 \specialrule{0em}{-1pt}{-1pt}
 & Dharmasiri \emph{et al.} \cite{8205954} & VGG &0.531& {\bf0.509} & 0.506\\
 \specialrule{0em}{-1pt}{-1pt}
 &  Laina \emph{et al.} \cite{7785097} &ResNet-50& 0.536 &0.422 & 0.463\\
 \specialrule{0em}{-1pt}{-1pt}
 &  Xu \emph{et al.} \cite{8099508} &ResNet-50& 0.600 & 0.366 & 0.439\\
 \specialrule{0em}{-1pt}{-1pt}
 & Fu \emph{et al.} \cite{DORN} &ResNet-101& 0.316 & 0.473 &0.412 \\
 \specialrule{0em}{-1pt}{-1pt}
 &  Hu \emph{et al.} \cite{revisit} &ResNet-50& 0.664& 0.476 &0.547 \\
 \specialrule{0em}{-1pt}{-1pt}
 & Lee \emph{et al.} \cite{Lee_2019_CVPR} & DenseNet-162 & 0.648 & 0.331 & 0.424 \\
 \specialrule{0em}{-1pt}{-1pt}
 & Chen \emph{et al.}* \cite{Chen2019SARPN} & ResNet-50 & \textbf{\color{red}0.675} &0.488 & \textbf{\color{red}0.559} \\
 \specialrule{0em}{-1pt}{-1pt}
 &  Ramamonjisoa \emph{et al.} \cite{Ramamonjisoa_2019_ICCV_Workshops} & ResNet-50 &0.598&0.338&0.419 \\
 \specialrule{0em}{-1pt}{-1pt}
 & Yin \emph{et al.} \cite{0Enforcing} & ResNeXt-101 &0.605&0.457&0.510\\
 \specialrule{0em}{-1pt}{-1pt}
 &  Ours & ResNet-50 & 0.665 &{\color{red}0.492} &0.558\\
 \specialrule{0em}{-1pt}{-1pt}
 \hline
 \specialrule{0em}{-1pt}{-1pt}
\multirow{12}{*}{\textgreater{}1} 
 & Eigen \emph{et al.} \cite{Eigen} & - &0.730&0.347 &0.456\\
 \specialrule{0em}{-1pt}{-1pt}
 & Eigen \emph{et al.} \cite{Fergus} &VGG& 0.733 & 0.488 & 0.574\\
 \specialrule{0em}{-1pt}{-1pt}
 & Dharmasiri \emph{et al.} \cite{8205954} & VGG & 0.617& 0.489&0.533\\
 \specialrule{0em}{-1pt}{-1pt}
 & Laina \emph{et al.} \cite{7785097} & ResNet-50 & 0.670 & 0.479 & 0.548 \\
 \specialrule{0em}{-1pt}{-1pt}
 &  Xu \emph{et al.} \cite{8099508} & ResNet-50 & 0.794 & 0.407 & 0.525\\
 \specialrule{0em}{-1pt}{-1pt}
 & Fu \emph{et al.} \cite{DORN} & ResNet-101 & 0.483 & 0.512 & 0.485 \\
 \specialrule{0em}{-1pt}{-1pt}
 & Hu \emph{et al.} \cite{revisit} & ResNet-50 & 0.755 &0.514 &0.604 \\
 \specialrule{0em}{-1pt}{-1pt}
 & Lee \emph{et al.} \cite{Lee_2019_CVPR} & DenseNet-162 & {\bf0.876} & 0.391 & 0.525 \\
 \specialrule{0em}{-1pt}{-1pt}
 & Chen \emph{et al.}* \cite{Chen2019SARPN} & ResNet-50 & 0.763 &0.526 & \textbf{\color{red}0.614} \\
 \specialrule{0em}{-1pt}{-1pt}
 &  Ramamonjisoa \emph{et al.} \cite{Ramamonjisoa_2019_ICCV_Workshops} & ResNet-50 &{\color{red}0.797}&0.404&0.524 \\
 \specialrule{0em}{-1pt}{-1pt}
 & Yin \emph{et al.} \cite{0Enforcing} & ResNeXt-101 &0.740&0.502&0.589\\
 \specialrule{0em}{-1pt}{-1pt}
 & Ours & ResNet-50 &0.750 & {\color{red}\bf0.531} & 0.613 \\

\hline
\end{tabular}
\end{center}
\caption{Depth boundary accuracy of different methods on the NYUD v2 dataset. The bold type indicates the best performance. The red number indicates the best performance with the same backbone of ResNet-50.
* Using a ResNet-50 backbone,
which is different from the original paper.}
\label{table:edge}
\end{table}

\subsubsection{Quantitative Evaluation}
Table \ref{table:depth} illustrates the comparisons between our proposed method and the recent approaches with ResNet-50 backbone.
The first three metrics (i.e., accuracy under threshold $t_d$($t_d\in[1.25,1.25^2,1.25^3]$)) measure the pixel-wise accuracy of the predicted depth map.
Note that,
SARPN \cite{Chen2019SARPN}\footnote{As SARPN employs the SENet as the backbone, in order to conduct the comparison objectively,
we modify its backbone to ResNet-50,
which is the same as our method.
Thus, the quantitative values of SARPN is not exactly equal to the original paper.}
employ the ResNet-50 as the backbone,
which is different from the official implementation.
In addition,
for other methods providing the source code,
the quantitative values are obtained by running their source code.
Since our platform may be slightly different from their original papers,
there are small differences in performance probably,
however, the differences are very small.
As a result,
our method can achieve the second best performance.
Another method \cite{Ramamonjisoa_2019_ICCV_Workshops} employs additional occlusion boundary label,
and this is the reason why their method achieves a higher accuracy than ours.
In addition, our method achieves a better accuracy than other methods \cite{Eigen,Fergus,8205954,7785097,revisit,8578175,8578140},
because the proposed network combines outputs of the PSE and long-distance contextual correlation,
which effectively reserves the contextual information with global depth layout.
Thus, the farthest region in the predicted depth is more accurate than before.
The latter three metrics (i.e., REL, RMS, log10) evaluate the mean pixel-wise error between the predicted depth and the true depth.
The result is similar to the first three metrics.
Although our method is not the best competing with some other methods based on extra labels \cite{8099508,Ramamonjisoa_2019_ICCV_Workshops}, heavy backbone \cite{0Enforcing,icip}, much more training data \cite{8578175}, or heavy structure \cite{Chen2019SARPN},
in general, our algorithm is better than the others.
The reason is that the SRM refines the depth in the absolute scale by exploiting a larger range of information and accurate locations of large depth changes.

As a significant metric for measuring the usability of the depth map,
the depth gradient around the boundary greatly affects the visualization of the generated depth.
Table \ref{table:edge} illustrates the accuracy of predicted boundary pixels in the depth maps.
Intuitively,
our method achieves a highest recall in recovering boundary with large depth change,
and is competitive with SARPN \cite{Chen2019SARPN} in the precision of depth boundary.
In the F1 score,
the proposed method is also competitive to SARPN \cite{Chen2019SARPN},
which also leads to good visualizations (see the next sub-section).
The performance is attributed to the combination of all modules.
The DCE provides context with a large receptive field to eliminate the false-positive pixels.
The BUBF extracts the boundary from the image.
By exploiting high-level features,
the BUBF refines the low-level feature gradually and preserves the boundary indicating depth change.
Guided by the gradient loss,
the BUBF eventually extracts the boundary which reveals the depth change.
In the SRM,
the feature maps with information of the boundary cue are used to refine the depth around the edge.

\begin{table}[!tb]
\footnotesize
\begin{center}
\begin{tabular}{|l|c|p{1.5cm}<{\centering}p{1.5cm}<{\centering}p{1.5cm}<{\centering}|}
\Xhline{1.2pt}
\multirow{2}{*}{Method and Variants}& \multirow{2}{*}{Backbone} & \multicolumn{3}{c|}{Error of the farthest region in predicted depth}\\
& & $m=6$ & $m=12$ & $m=24$\\
\hline
\hline
Eigen \emph{et al.} \cite{Eigen} & - & 0.1692 & 0.1827 & 0.1871 \\
Fergus \emph{et al.} \cite{Fergus}&VGG&0.1181 & 0.1354 &0.1460 \\
Dharmasiri \emph{et al.} \cite{8205954} & VGG &0.1573 &0.1725 & 0.1796\\
Laina \emph{et al.} \cite{7785097}&ResNet-50&0.1162 &0.1400&0.1573 \\
Fu \emph{et al.} \cite{DORN} & ResNet-101 &0.1090 & 0.1254 & 0.1281 \\
Hu \emph{et al.} \cite{revisit}& ResNet-50&0.1123 	&0.1308 &0.1428 \\
Lee \emph{et al.} \cite{Lee_2019_CVPR}&DenseNet-162 &0.1214 &0.1408 &0.1496 
\\
Chen \emph{et al.}* \cite{Chen2019SARPN} &ResNet-50 &\textcolor{red}{0.1029} & \textcolor{red}{0.1225} &0.1350 \\
 Ramamonjisoa \emph{et al.} \cite{Ramamonjisoa_2019_ICCV_Workshops} & ResNet-50 &0.1162 &0.1352 &0.1475 
\\
Yin \emph{et al.} \cite{0Enforcing} & ResNeXt-101 &\textbf{0.0880} &\textbf{0.1018} &\textbf{0.1143} \\
\hline
Baseline &ResNet-50& 0.1113 &0.1338 & 0.1427 \\
Baseline + DCE &ResNet-50& 0.1122 & 0.1314 & 0.1402\\
Baseline + BUBF + SRM &ResNet-50& 0.1085 & 0.1279 &0.1373 \\
Ours &ResNet-50& 0.1061 & 0.1263 & \textcolor{red}{0.1349} \\
\Xhline{1.2pt}
\end{tabular}
\end{center}
\caption{Farthest region accuracy under partition ratios of different methods on NYUD v2 dataset. The bold type indicates the best performance. The red number indicates the best performance with the same backbone of ResNet-50.
* Using a ResNet-50 backbone,
which is different from the original paper.}
\label{table:vp}
\end{table}

Table 3 shows the performance comparison of different methods on the proposed distance error of the farthest region.
It can be seen that,
compared to methods using the same backbone,
our method achieves similar performance with SARPN \cite{Chen2019SARPN}.
They even surpass several methods using a heavy backbone and relative depth regression.
In addition,
the variant Baseline+BUBF+SRM reduces the error of farthest region in all scales,
because the depth boundary near the farthest region helps to locate the farthest region.
The variant Baseline+DCE also outperforms the baseline.
The reason is that our method obtains long-distance context correlation through DCE.
When using all proposed modules,
our method outperforms almost methods with the same backbone,
even is competitive with some methods with a heavier backbone.

\subsubsection{Qualitative Evaluation}
\begin{figure}
	\centering
	\includegraphics[width=1\linewidth]{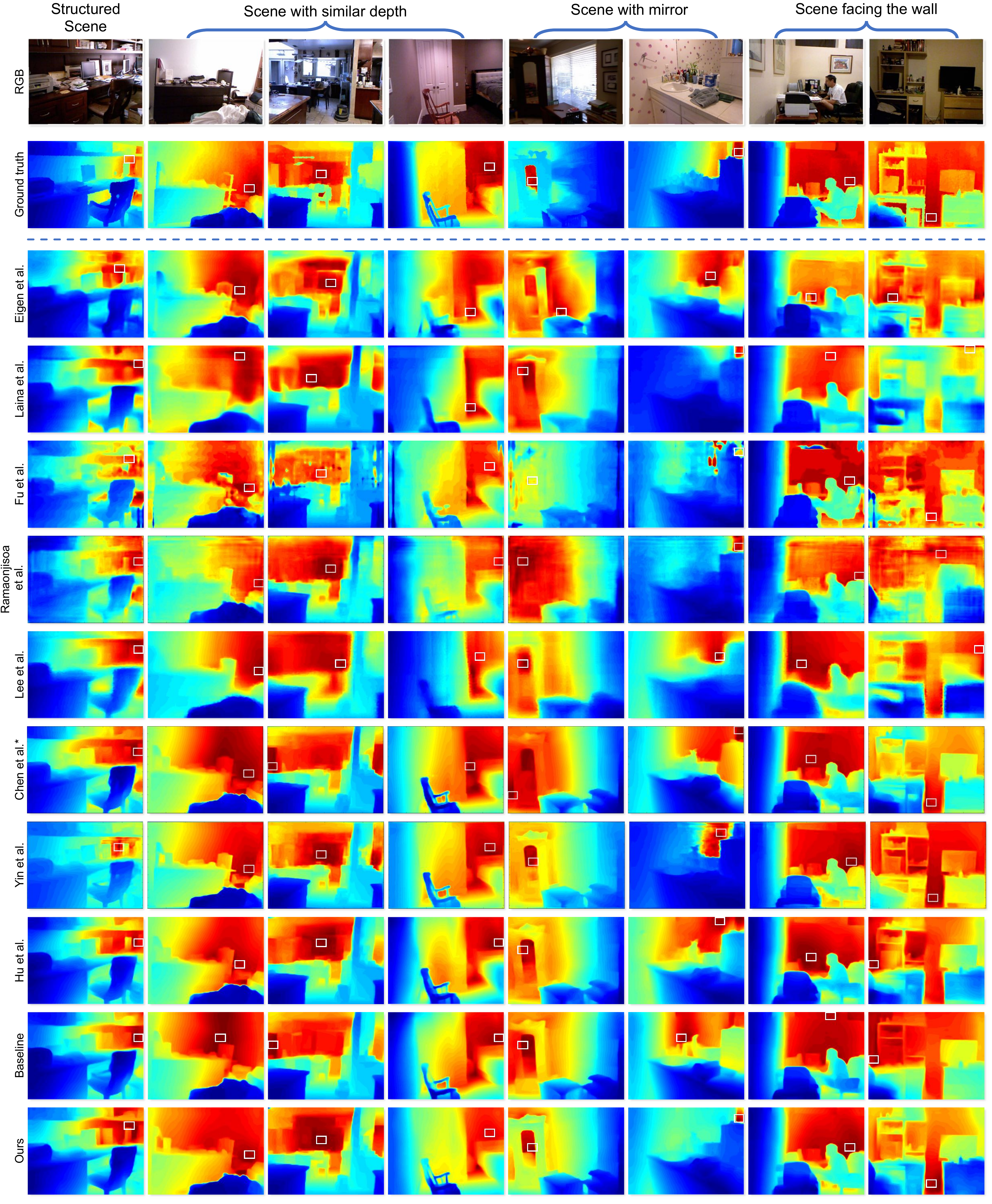}
	\caption{
	Example depth maps predicted by several methods and our proposed method.
	$1^{st}$ row: input images;
	$2^{nd}$ row: ground truth;
	$3^{rd}-12^{th}$ rows: predicted depth maps of Eigen et at. \cite{Fergus}, Liana \emph{et al.} \cite{7785097}, Fu \emph{et al.} \cite{DORN}, Ramamonjisoa \emph{et al.} \cite{Ramamonjisoa_2019_ICCV_Workshops}, Lee \emph{et al.} \cite{Lee_2019_CVPR}, Chen \emph{et al.}* \cite{Chen2019SARPN}, Yin \emph{et al.} \cite{0Enforcing}, Hu \emph{et al.} \cite{revisit}, our baseline and the proposed method.
	The white boxes mark the predicted farthest region ($m=12$).
* Using a ResNet-50 backbone,
which is different from the original paper.}
	\label{fig:quality}
\end{figure}

Figure.\ref{fig:quality} shows the qualitative results on the NYUD v2 dataset.
The first row to the twelfth row shows the original RGB images,
ground truth,
the depth maps predicted by \cite{Fergus,7785097,DORN,Ramamonjisoa_2019_ICCV_Workshops,Lee_2019_CVPR,Chen2019SARPN,0Enforcing,revisit},
our baseline, and the proposed method,
respectively.
The depth maps are visualized by different colors corresponding to different depth values,
i.e., dark blue corresponds to the minimum depth and dark red corresponds to the maximum depth.
The white rectangular boxes mark the farthest region in the depth maps.

\begin{table}[!tb]
\footnotesize
\begin{center}
\begin{tabular}{|l|p{0.8cm}<{\centering}p{0.8cm}<{\centering}p{0.8cm}<{\centering}|p{0.8cm}<{\centering}p{0.8cm}<{\centering}p{0.8cm}<{\centering}|}
\Xhline{1.2pt}
\multirow{2}{*}{Var}& \multicolumn{3}{c|}{higher is better ($\delta$\textless)} & \multicolumn{3}{c|}{lower is better}\\
 & $1.25$ & $1.25^{2}$ & $1.25^{3}$ & RMS & REL & log10\\
\hline
\hline
Baseline & 0.840 & 0.966 &0.991&0.557 &0.128 &0.055 \\
+DCE &\textbf{0.848}	&0.967	&0.991&0.555&0.125	&0.0537	 \\
+BUBF+SRM &0.838	&0.969	&0.992&0.560	&0.126	&0.055	\\
+DCE+BUBF+SRM &0.846 &\textbf{0.969}&\textbf{0.992}& \textbf{0.550}&\textbf{0.123}&\textbf{0.053} \\
\Xhline{1.2pt}
\end{tabular}
\end{center}
\caption{The prediction result on NYUD v2 dataset.
The bold type indicates the best performance.}
\label{table:abla_1_depth}
\end{table}

\begin{figure}[!tb]
	\centering
	\includegraphics[width=1\linewidth]{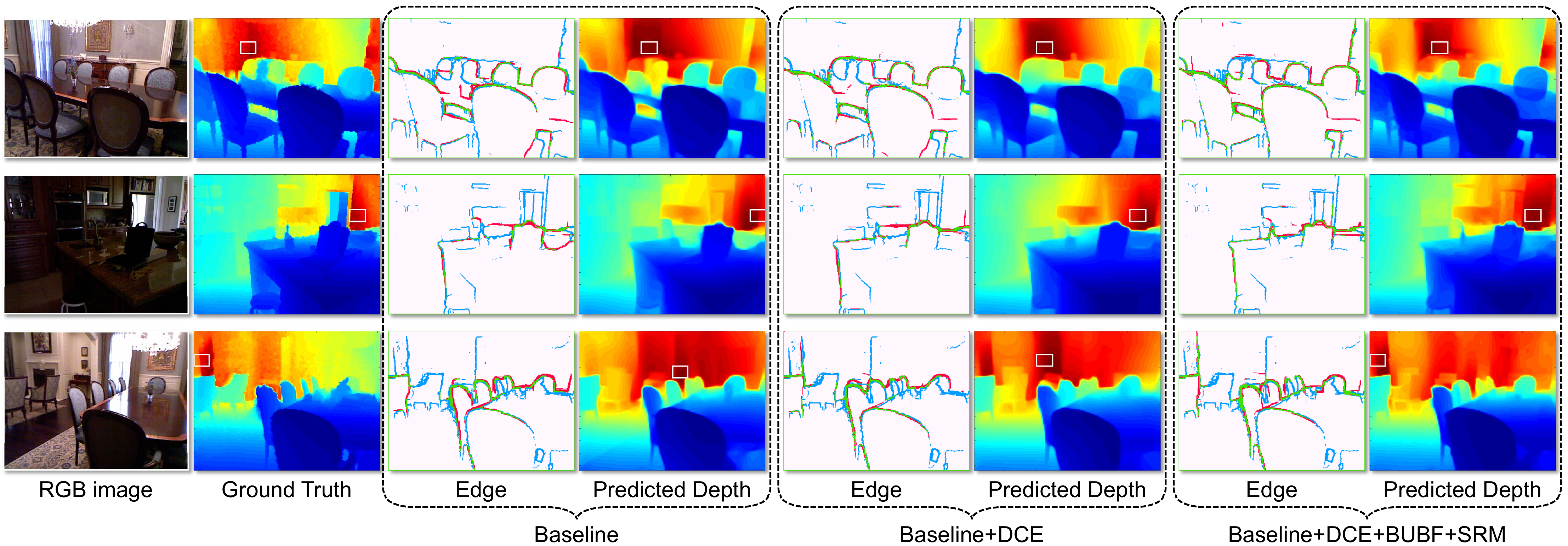}
	\caption{
	Example predicted depth and visualized boundary of several variants.
	Following the boundary metric,
	the sobel operation is used to recover boundary from depth map,
	and pixels larger than 1 are regarded as boundary pixels.
	Green indicates true positive pixels,
	red indicates false positive pixels,
	and blue indicates false negative pixels.}
	\label{fig:ablation1}
\end{figure}

In detail,
the first column shows an office desk and a chair with a clear depth.
Due to numerous objects stacking on the desk,
it is hard to finely recover the depth changes between these objects.
Obviously,
other methods predict the wrong farthest region and blurred depth changes,
e.g., the monitor on the desk.
In contrast,
our method fully predicts the real depth of each object without any blur
and correctly recovers the farthest region,
which reveals the effectiveness of boundary cues and extracting global information.
The scenes in the second to fourth columns show a similar depth over a large area,
e.g., the wall in these scenes.
Due to the lack of global context,
other methods fail to correctly find the farthest region in these scenes
and thus suffer from the distortion of the 3D scene structure.
But our method works admirably in predicting depth in these scenes.
Intuitively,
comparing to others,
our method avoids the grids artificial on the wall
and completely recovers the shape detail of objects,
especially the wooden chair in the fourth column.
The reason is that the boundary cue better locates the depth change than other edge cues
and the global context is necessary.
In the fifth and sixth columns,
due to the mirror in the scene,
it is hard to predict the depth change between areas inside and outside the mirror.
Most methods fail to predict the depth level between the two areas,
causing the distorted 3D scene structure.
By utilizing the boundary cue,
our method fully predicts the depth level of the scene.
In the last two columns,
since the depths of the wall facing the camera are extremely similar,
it is very difficult to accurately predict its depth.
It can be seen that all
most other methods fail to predict the farthest region.
By utilizing the global context of the scene and the boundary cue,
our method correctly predicts the depth.

\subsubsection{Ablation Studies}
To clarify the contribution of each proposed module,
the baseline is defined as the combination of our encoder and decoder,
and several ablation studies are conducted to verify the improvement.

\begin{table}[!tb]
\footnotesize
\begin{center}
\begin{tabular}{|l|l|ccc|}
\hline
Thres & Method & P & R & F1 \\
\hline
\multirow{4}{*}{\textgreater{}0.25} 
 & +DCE & 0.628  & 0.457 & 0.523 \\
 & +BUBF+SRM  & 0.636 &0.489&0.547\\
 & +DCE+BUBF+SRM & \textbf{0.645}  & \textbf{0.489} & \textbf{0.550} \\
 \hline
\multirow{4}{*}{\textgreater{}0.5}
 & +DCE & 0.648 &0.458 & 0.530 \\
 & +BUBF+SRM  &0.638&0.488&0.547\\
 & +DCE+BUBF+SRM & \textbf{0.665} & \textbf{0.492} & \textbf{0.558} \\
 \hline
 \multirow{4}{*}{\textgreater{}1}
 & +DCE & 0.733 &0.507 & 0.592 \\
 & +BUBF+SRM  & 0.638&0.489& 0.547\\
 & +DCE+BUBF+SRM & \textbf{0.750} &\textbf{0.531} &\textbf{0.613} \\
\hline
\end{tabular}
\end{center}
\caption{Accuracy of predicted boundary pixels in depth maps under different thresholds.
The bold type indicates the best performance.}
\label{table:abla_1_edge}
\end{table}

{\bf{Role of the} global context and boundary cue:}
Two issues are assumed in this paper,
i.e., (1) global context effectively improves the prediction of depth layout.
(2) boundary cue boosts the prediction of the depth gradient around the object boundary.
To verify them,
three variants are constructed for evaluation,
as shown in Table \ref{table:abla_1_depth} and Table \ref{table:abla_1_edge}.
Intuitively,
compared to the baseline,
since DCE extracts the global contextual information to predict the farthest region of the scene,
the variant Baseline+DCE achieves better performance in all metrics of evaluation.
By combining the baseline, the BUBF, and the SRM,
the variant Baseline+BUBF+SRM locates the edge with depth change more precisely
and fits the depth near the contours more precisely.
Note that,
in Table \ref{table:abla_1_edge},
DCE seems to play a more significant role than BUBF + SRM in the depth boundary recovery when the threshold is larger than 1.
Since the boundary is not a local visual cue,
the network needs sufficient contextual cues to suppress boundary that are visually significant but have no depth changes.
It makes the baseline + DCE to eliminate many false-positive boundary pixels,
even outperform the variant baseline + BUBF + SRM in depth boundary prediction.
Furthermore,
the whole network with all modules improves the accuracy of pixel-wise depth while ensuring the sharp detail of objects.
The intrinsic reason is that the DCE and BUBF play important and different roles in depth prediction.
Specifically,
on the one hand,
the DCE aggregates the global information with a larger perceptive field,
boosting the accuracy in global depth layout prediction.
On the other hand,
the combination of BUBF and SRM extracts the boundary cue
and then refines the depth change around the boundary.

Fig.\ref{fig:ablation1} illustrates the visualized results of several variants.
Compared to the baseline,
the other two variants generate fewer false-positive boundary pixels (marked in red),
such as the surface of the desk in the first row,
the small objects on the desk of the second row,
and the chairs boundary in the third row.
Among them,
the variants Baseline+DCE+BUBF+SRM performs the best.
The reason is that,
BUBF and SRM improve the accuracy of the overall boundary.
In addition,
the two variants both achieve a lower distance error of the farthest region than the baseline.

{\bf{Contribution of the} Pyramid Scene Encoder:}
The PSE applies average pooling with various kernel sizes,
then integrates scene features from different sub-regions.
The global contextual information in sub-regions helps to estimate the depth layout effectively.
Table \ref{table:abla_2_pooling} indicates various experimental results with the different settings of pooling kernel size in the PSE.
We first remove all pooling layers,
and then gradually add them back.
Intuitively,
these experiments prove that the setting,
in which the sizes of feature maps after pooling are 1,2,3 and 6 respectively,
achieves the best performance.
It illustrates that different from the multi-scale dilated convolutions,
the PSE aggregates the global information which is significant for predicting the global depth layout.

\begin{table}[!tb]
\footnotesize
\begin{center}
\begin{tabular}{|l|p{0.6cm}<{\centering}p{0.6cm}<{\centering}p{0.6cm}<{\centering}|p{0.6cm}<{\centering}p{0.6cm}<{\centering}p{0.6cm}<{\centering}|}
\Xhline{1.2pt}
\multirow{2}{*}{\tabincell{c}{pooling\\$i \times i$}}& \multicolumn{3}{c|}{higher is better ($\delta$\textless)} & \multicolumn{3}{c|}{lower is better}\\
 & $1.25$ & $1.25^{2}$ & $1.25^{3}$ & RMS & REL & log10\\
\hline
\hline
None &0.844 &0.967 &0.991 &0.553 & 0.125 &0.053\\
\emph{i$\in$\{1\}}	&0.843 &0.967 &0.991 &0.562 &0.125 &0.054\\
\emph{i$\in$\{1,2\}} 	&0.844	&0.967	&0.991&0.560	&0.125	&0.054\\
\emph{i$\in$\{1,2,3\}} &0.842&0.967&0.991&0.560 &0.126 &0.054\\
\emph{i$\in$\{1,2,3,6\}} & \textbf{0.846}& \textbf{0.969} & \textbf{0.992} & \textbf{0.550} & \textbf{0.123} & \textbf{0.053}\\
\Xhline{1.2pt}
\end{tabular}
\end{center}
\caption{The experimental result of pyramid scene encoder with diffierent pooling rate.
The bold type indicates the best performance.}

\label{table:abla_2_pooling}
\end{table}
\begin{figure}[!tb]
\centering
\includegraphics[width=1\linewidth]{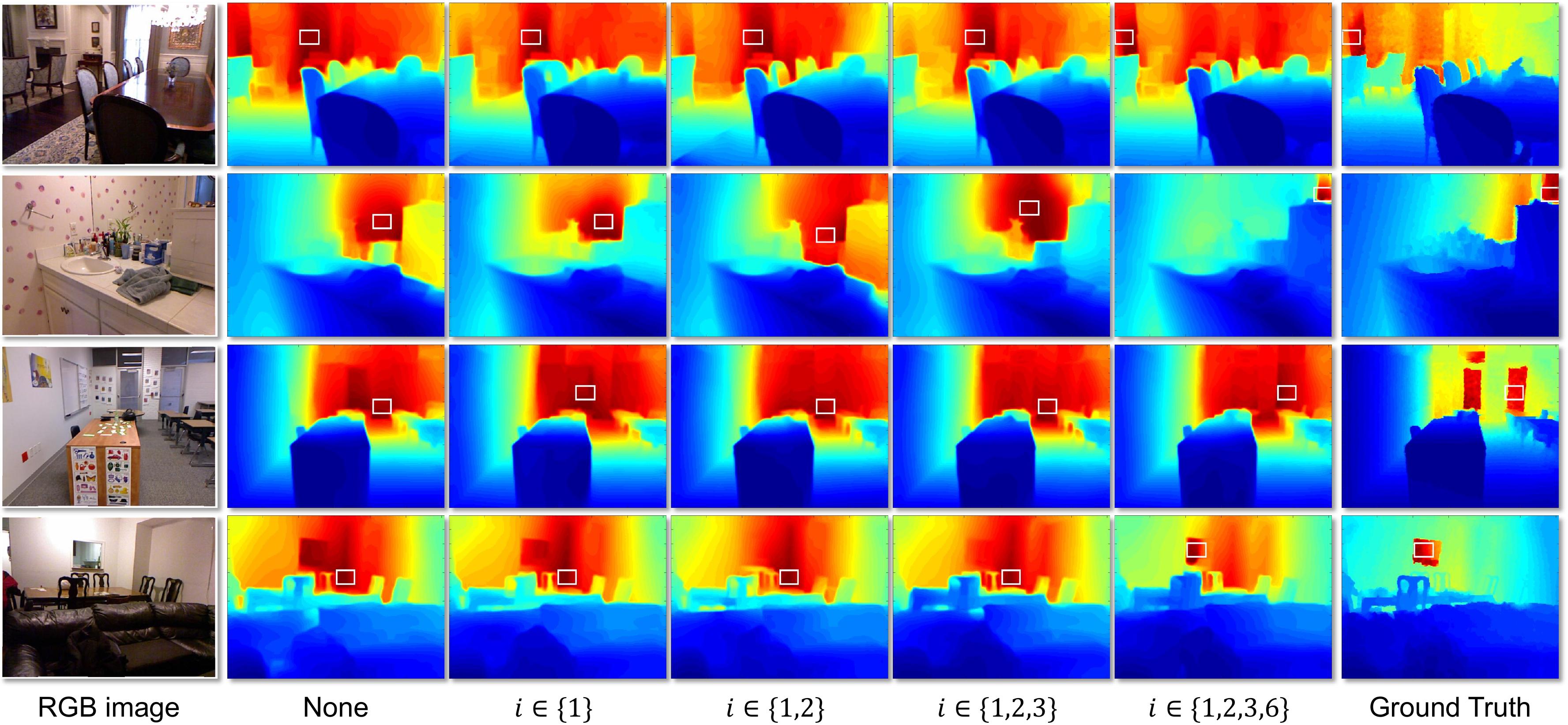}
\caption{Example predicted depth of several variants with different pooling rates.
The white boxes indicate the farthest regions.}
\label{fig:ablation2}
\end{figure}

Fig. \ref{fig:ablation2} illustrates the visualization results of PSE variants.
It can be seen that the predicted depth is getting closer to the real depth as more pooling layers are added.
In the first row,
as more pooling layers are used,
the network gradually predicts the farthest region in the scene.
In the second row,
when more pooling layers are used,
the farthest region in the mirror is located more accurately.
In the third row,
it is difficult to determine which area on either side of the pillar is further.
As more pooling layers are used,
the farthest region is accurately predicted.
In the fourth row,
the hollow on the wall can easily be regarded as coplanar with the wall.
When the $6\times 6$ pooling layer is employed,
the farthest region in this hollow is accurately predicted.

\begin{table}[]
\footnotesize
\begin{center}
\begin{tabular}{|l|l|ccc|}
\hline
Thres & Method & Precision & Recall & F1-score \\
\hline
\hline
\multirow{3}{*}{\textgreater{}0.25} &Baseline & 0.629 & 0.457 & 0.523\\
 & Ours/with MFF \cite{revisit} & 0.638 & 0.483 & 0.550 \\
 & Ours/with BUBF & \textbf{0.645} & \textbf{0.489} & \textbf{0.550} \\
 \hline
\multirow{3}{*}{\textgreater{}0.5} & Baseline & 0.648 & 0.458 & 0. 530 \\
 & Ours/with MFF \cite{revisit} & 0.659 & \textbf{0.493} & 0.557 \\
 & Ours/with BUBF & \textbf{0.665} & 0.492 & \textbf{0.558} \\
 \hline
\multirow{3}{*}{\textgreater{}1} & Baseline & 0.733 & 0.507 & 0.592 \\
 & Ours/with MFF \cite{revisit} & 0.745 & 0.531 & 0.612 \\
 & Ours/with BUBF & \textbf{0.750} & \textbf{0.531} & \textbf{0.613 } \\
\hline
\end{tabular}
\end{center}
\caption{Accuracy of predicted boundary pixels in depth maps under different thresholds.
The bold type indicates the best performance.}
\label{table:abla_2_OL}
\end{table}

\begin{figure}[!tb]
	\centering
	\includegraphics[width=1\linewidth]{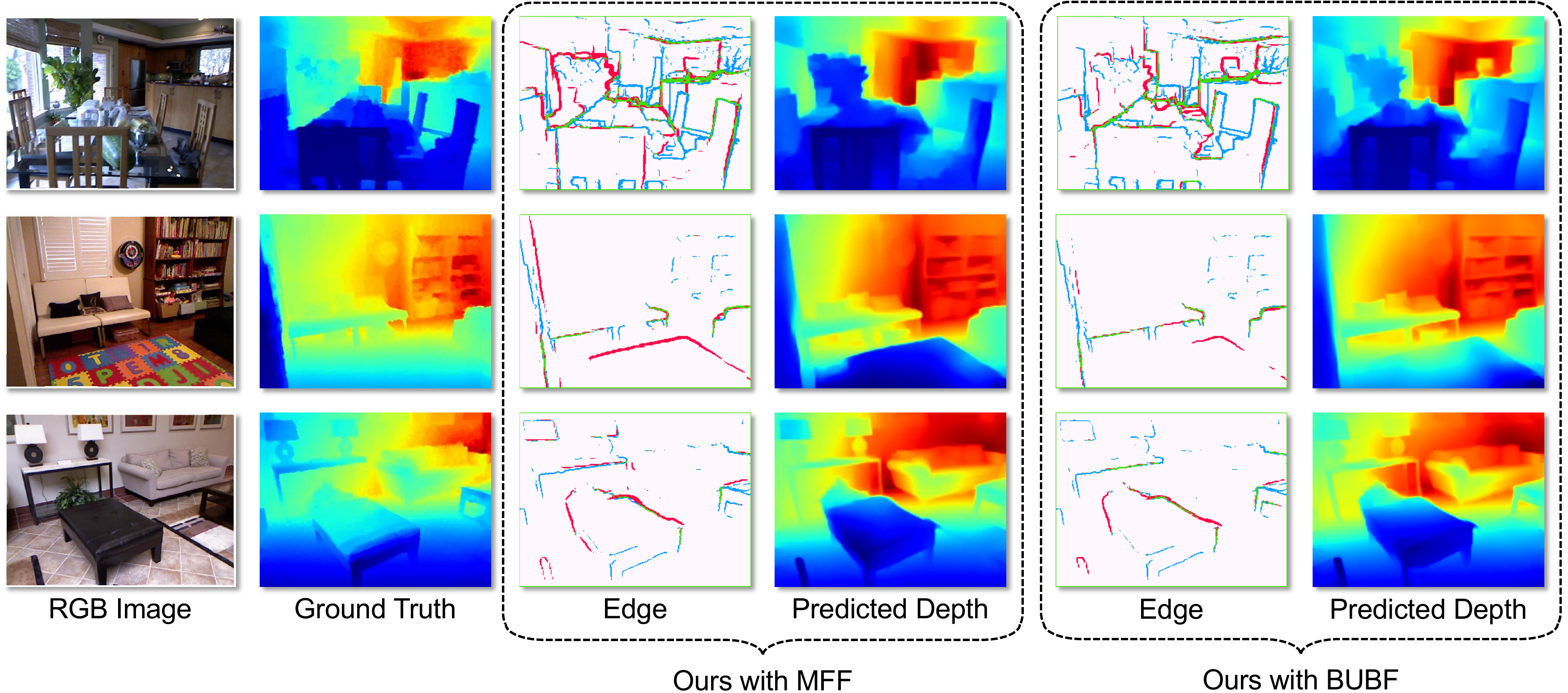}
	\caption{Example predicted depth and visualized boundary of several variants.
	Following the boundary metric,
	the sobel operation is used to recover boundary from depth map,
	and pixels larger than 1 are regarded as boundary pixels.
	Green indicates true positive pixels,
	red indicates false positive pixels,
	and blue indicates false negative pixels.}
	\label{fig:ablation3}
\end{figure}

{\bf{Learning method for boundary cue:}}
There are previous methods that apply the multi-level information from encoder to improve the edge accuracy,
such as multi-scale feature fusion (MFF) \cite{Chen2019SARPN,revisit}.
Different from these methods,
the proposed BUBF finely learns the boundary cue from the scene,
i.e., the location with a sudden change in depth.
To evaluate the effectiveness of learning boundary,
the MFF is employed to compare with the proposed BUBF.
As shown in Table \ref{table:abla_2_OL},
our module BUBF better locates the object contour
and fine-tunes the accurate depth near the contour.
The reason is that the MFF has a problem of introducing lots of useless noisy cues in depth prediction,
which is addressed by our BUBF.

Furthermore, Fig. \ref{fig:ablation3} visualizes the results of our model with MFF and BUBF.
In the first row,
there are a large number of boundary pixels of the window and the plant.
Our method not only obtains fewer false-positive boundary pixels (marked in red),
but also obtains more true-positive boundary pixels (marked in green).
In the second row, 
there is a large carpet with various appearances.
Due to directly introducing low-level cues,
MFF leads to a large number of false-positive boundary pixels (marked in red),
while the proposed BUBF alleviates this issue better.
In the third row,
MFF locates the boundary of the desk inaccurately,
but BUBF locates these boundary pixels better.

\begin{table}[!tb]
\footnotesize
\begin{center}
\begin{tabular}{|c|c|ccc|ccc|}
\Xhline{1.2pt}
\multirow{2}{*}{Method}&\multirow{2}{*}{backbone}&\multicolumn{3}{c|}{higher is better ($\delta$\textless)}& \multicolumn{3}{c|}{lower is better}\\
 & & $1.25$ & $1.25^2$ & $1.25^3$ & REL & RMS & log10 \\
\hline
\hline
Eigen \emph{et al.} \cite{Eigen} & - & 0.36& 0.65& 0.84 & 0.32 & 1.55 & 0.17\\
Eigen \emph{et al.} \cite{Fergus} & AlexNet & 0.40 & 0.73 & 0.88& 0.30 & 1.38 & 0.15 \\
Eigen \emph{et al.} \cite{Fergus} & VGG & 0.47 & 0.78 & 0.93& 0.25 & 1.26 & 0.13 \\
Dharmasiri \emph{et al.} \cite{8205954} & VGG &0.22 &0.55&0.78&0.35& 1.61 &0.19\\
 Laina \emph{et al.} \cite{7785097} & ResNet-50 & 0.50 & 0.78 & 0.91& 0.26 & 1.20 & 0.13 \\
Hu \emph{et al.} \cite{revisit}& ResNet-50&0.52	&\textcolor{red}{0.85}&\textcolor{red}{\textbf{0.95}} &0.23 &1.13&0.12\\
Fu \emph{et al.} \cite{DORN} & ResNet-101 &0.55&0.81&0.92&0.24&1.13&0.12 \\
Lee \emph{et al.} \cite{Lee_2019_CVPR} & DenseNet-162 & 0.53&0.83&0.95&0.23&1.09&0.11 \\
Chen \emph{et al.}* \cite{Chen2019SARPN}& ResNet-50 &0.51&0.84&0.94&\textcolor{red}{0.22}&1.14&\textcolor{red}{0.11} \\
Liu \emph{et al.} \cite{2015Deep} & - &0.48 & 0.78 & 0.91& 0.30 & 1.26 & 0.13 \\
Li \emph{et al.} \cite{Li_iccv} & VGG &0.58 & 0.85 & 0.94& 0.22 &1.09 & 0.11 \\
Liu \emph{et al.} \cite{Liu_cvpr} & - &0.41 & 0.70 & 0.86& 0.29 & 1.45 & 0.17 \\
 Ramamonjisoa \emph{et al.}
 \cite{Ramamonjisoa_2019_ICCV_Workshops} & ResNet-50 &\textcolor{red}{ 0.59} & 0.84&0.94& 0.26 & \textcolor{red}{1.07} & \textcolor{red}{0.11}\\
Yin \emph{et al.} \cite{0Enforcing} & ResNeXt-101 &0.54&0.84&0.93&0.24&1.06&0.11\\
Swami \emph{et al.} \cite{icip} & SENet-154 &\textbf{0.60}&\textbf{0.87}&\textbf{0.95}&\textbf{0.20}&\textbf{1.03}&\textbf{0.10} \\
\hline
 \textbf{Ours} & ResNet-50 & 0.51&0.82&0.93& 0.24 & 1.19 & 0.12\\
\Xhline{1.2pt}
\end{tabular}
\end{center}
\caption{Conventional depth error and accuracy on iBims-1 dataset. The bold type indicates the best performance. The red number indicates the best performance with the same backbone of ResNet-50.
* Using a ResNet-50 backbone,
which is different from the original paper.}
\label{table:depth_ibims1}
\end{table}

\begin{table}[!tb]
\footnotesize
\begin{center}
\begin{tabular}{|p{2.4cm}<{\centering}|c|cc|cc|ccc|}
\Xhline{1.2pt}
\multirow{2}{*}{Method}&\multirow{2}{*}{backbone}&\multicolumn{2}{c|}{PE (cm$/^{\circ}$)$\downarrow$}&\multicolumn{2}{c|}{DBE (px)$\downarrow$}& \multicolumn{3}{c|}{DDE (\%)}\\
 & & $\epsilon^{plan}$ & $\epsilon^{orie}$ & $\epsilon^{acc}$ & $\epsilon^{comp}$ & $\epsilon^{0}$$\uparrow$ & $\epsilon^{+}$$\downarrow$ & $\epsilon^{-}$$\downarrow$ \\
\hline
\hline
Eigen \emph{et al.} \cite{Eigen} & -&7.70&24.91 & 9.97& 9.99& 70.37& 27.42& 2.22\\
Eigen \emph{et al.} \cite{Fergus} & AlexNet &7.52&21.50 & 4.66 & 8.68 & 77.48 & 18.93 & 3.59 \\
Eigen \emph{et al.} \cite{Fergus} & VGG&5.97& 17.65 & 4.05 & 8.01 & 79.88 & 18.72 & 1.41 \\
Dharmasiri \emph{et al.} \cite{8205954} & VGG &6.97 & 28.56 & 5.07& 7.83 & 70.10&29.46&0.43 \\
 Laina \emph{et al.} \cite{7785097} & ResNet-50 &6.46& \textcolor{red}{19.13} & 6.19 & 9.17 & 81.02 & 17.01 & 1.97 \\
Hu \emph{et al.} \cite{revisit}& ResNet-50&3.88	&28.06&2.36&5.40 &82.20&16.10&1.69\\
Fu \emph{et al.} \cite{DORN} &ResNet-101&10.50&23.83 &4.07&-&82.78&-&- \\
Lee \emph{et al.} \cite{Lee_2019_CVPR} &DenseNet-162& \textbf{2.27}&20.73&6.21&8.66&84.35&12.72&2.92 \\
Chen \emph{et al.}* \cite{Chen2019SARPN} & ResNet-50 &\textcolor{red}{3.45}&43.44&2.98&\textbf{\textcolor{red}{4.96}}&82.27&16.38&\textcolor{red}{1.34} \\
Liu \emph{et al.} \cite{2015Deep} & -& 8.45 & 28.69 &2.42 & 7.11 & 79.70 & 14.16 & 6.14 \\
Li \emph{et al.} \cite{Li_iccv} & VGG &7.82& 22.20 &3.90 & 8.17 & 83.71 & 13.20 & 3.09 \\
Liu \emph{et al.} \cite{Liu_cvpr} & -& 7.26& 17.24 &4.84 & 8.86 & 71.24 & 28.36 & \textbf{0.40} \\
 Ramamonjisoa \emph{et al.} \cite{Ramamonjisoa_2019_ICCV_Workshops} & ResNet-50 &9.95& 25.67& 3.52 & 7.61 & \textcolor{red}{84.03} & \textbf{\textcolor{red}{9.48}} & 6.49 \\
Yin \emph{et al.} \cite{0Enforcing} & ResNeXt-101 &5.73&16.91&3.65&7.16&82.72&13.91&3.36 \\
Swami \emph{et al.} \cite{icip} & SENet-154 & 6.67&\textbf{16.52}&\textbf{2.06}&-&\textbf{84.96}&-&-\\
\hline
 \textbf{Ours} & ResNet-50 &3.98& 28.75 &\textcolor{red}{2.25} &5.18 & 80.54&17.64 &1.80 \\
\Xhline{1.2pt}
\end{tabular}
\end{center}
\caption{Planarity error, depth boundary errors, and directed depth error on iBims-1 dataset. The bold type indicates the best performance. The red number indicates the best performance with the same backbone of ResNet-50.
* Using a ResNet-50 backbone,
which is different from the original paper.}
\label{table:depth_ibims2}
\end{table}

\subsection{iBims-1 dataset}
iBims-1 is a new RGB-D dataset,
which contains pairs of a high-quality depth map and a high-resolution image.
These pairs are acquired by a digital single-lens reflex (DSLR) camera and a high-precision laser scanner.
Thus,
compared with NYUD v2 dataset,
iBims-1 achieves a very low noise level,
sharp depth transitions,
no occlusions,
and high depth ranges.
This dataset contains 100 pairs for evaluation,
and is also useful to evaluate the generalization of our method.
Note that,
since the iBims-1 dataset lacks training set,
all models used for testing on the iBims-1 dataset are trained on the NYUD v2 dataset.

\begin{table}[!tb]
\footnotesize
\begin{center}
\begin{tabular}{|l|c|p{1.5cm}<{\centering}p{1.5cm}<{\centering}p{1.5cm}<{\centering}|}
\Xhline{1.2pt}
\multirow{2}{*}{Method and Variants}& \multirow{2}{*}{Backbone} & \multicolumn{3}{c|}{Error of the farthest region in predicted depth}\\
& & $m=6$ & $m=12$ & $m=24$\\
\hline
\hline
Dharmasiri \emph{et al.} \cite{8205954} & VGG &0.1927&0.2011 & 0.2128\\
 Ramamonjisoa \emph{et al.} \cite{Ramamonjisoa_2019_ICCV_Workshops} & ResNet-50 &0.2020&0.2217&0.2400\\
Lee \emph{et al.} \cite{Lee_2019_CVPR} & DenseNet-162 & 0.1819&0.1961&0.2152\\
Laina \emph{et al.} \cite{7785097}&ResNet-50& \textcolor{red}{\textbf{0.1689}}&0.1916&0.2100 \\
Hu \emph{et al.} \cite{revisit}& ResNet-50&0.1804&0.2075&0.2147 \\
Chen \emph{et al.}* \cite{Chen2019SARPN} &ResNet-50 &0.1693& 0.1895 &0.2021 \\
Yin \emph{et al.} \cite{0Enforcing} & ResNeXt-101 &0.2022 &0.2086 &0.2343 \\
\hline
Ours &ResNet-50&0.1724 &\textcolor{red}{\textbf{0.1863}} &\textcolor{red}{\textbf{0.1981}}\\
\Xhline{1.2pt}
\end{tabular}
\end{center}
\caption{Distance error of the farthest region under different partition rate on iBims-1 datasets. The bold type indicates the best performance. The red number indicates the best performance with the same backbone of ResNet-50.
* Using a ResNet-50 backbone,
which is different from the original paper.}
\label{table:vp_ibims}
\end{table}

\begin{figure}[!tb]
	\centering
	\includegraphics[width=1\linewidth]{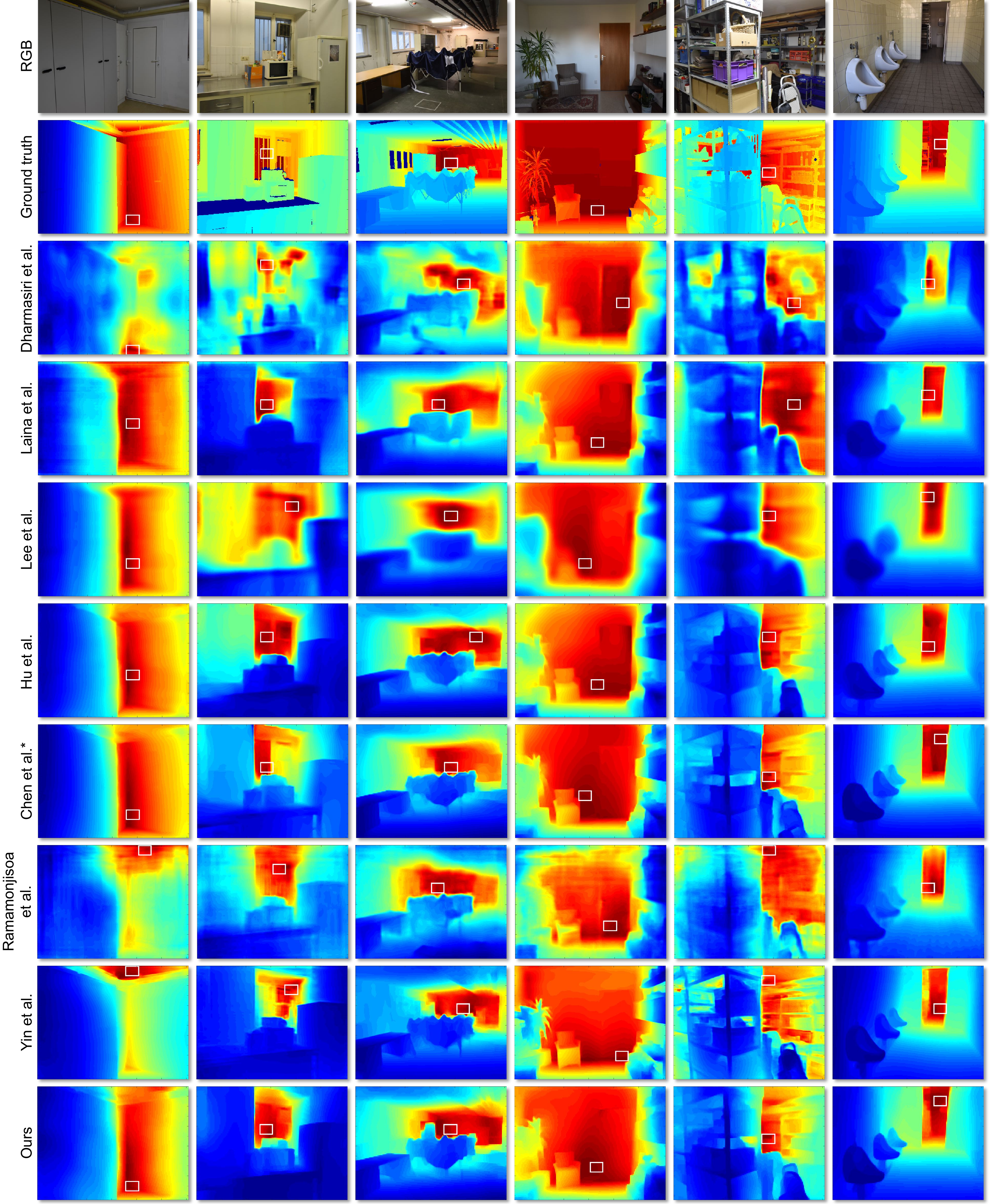}
	\caption{The visualization results on iBims-1 dataset.
	$1^{st}$ row: input images;
	$2^{nd}$ row: ground truth;
	$3^{rd}-10^{th}$ rows: predicted depth maps of Dharmasiri \emph{et at.} \cite{8205954}, Liana \emph{et al.} \cite{7785097}, Lee \emph{et al.} \cite{Lee_2019_CVPR}, Hu \emph{et al.} \cite{revisit}, Chen \emph{et al.}* \cite{Chen2019SARPN}, Ramamonjisoa \emph{et al.} \cite{Ramamonjisoa_2019_ICCV_Workshops}, Yin \emph{et al.} \cite{0Enforcing}, the proposed method.
	The white rectangular boxes mark the farthest region in the corresponding depth maps.
* Using a ResNet-50 backbone,
which is different from the original paper.}

	\label{fig:qualitative_ibims}
\end{figure}

\subsubsection{Quantitative Evaluation}
Table \ref{table:depth_ibims1} shows the performance of the commonly used error metrics on the iBims-1 dataset.
When $\delta$ is small than $1.25^2$ or $1.25^3$,
although our method does not focus on the overall depth accuracy and error,
it is still comparable with the previous methods.
Table \ref{table:depth_ibims2} illustrates the new metrics of different methods on the iBims-1 dataset.
It can be seen that our method achieves a low 3D planarity error.
The reason is that our method mainly focuses on the accurate estimation of the farthest region to improve the depth.
Noteworthy,
our method outperforms most methods on the depth boundary errors,
even outperforms the methods trained with extra boundary label \cite{Ramamonjisoa_2019_ICCV_Workshops} and some methods with a heavier backbone,
which proves the effectiveness of our method on improving the boundary prediction.
Table \ref{table:vp_ibims} illustrates the farthest region distance error of several methods on the iBims-1 dataset.
The method \cite{0Enforcing},
which performs well on NYUD v2 dataset,
however have a high error on the iBims-1 dataset.
Our method not only performs well on NYUD v2 dataset,
but also outperforms other methods on the iBims-1 dataset.
Our method achieves the lowest error when $m = 12$ and $m = 24$,
indicating that our method can accurately predict the farthest region.
Moreover, it also proves the generalization of our method.

\subsubsection{Qualitative Evaluation}
Fig. \ref{fig:qualitative_ibims} illustrates the visualized results of different methods on the iBims-1 dataset.
The first column shows a standard indoor scene.
Both our method and \cite{Chen2019SARPN} can correctly predict the farthest region.
The method \cite{0Enforcing} using the plane normal vector fails to correctly estimate the structure of the scene.
The second column shows a kitchen scene.
Most methods are hard to recover the farthest region of the scene (i.e., the window) and the shape details of the objects in the scene (the boundary of the window) at the same time,
but our method performs very well in both aspects.
The third column shows an office scene with an irregular tent placed inside.
Intuitively, the proposed method predicts the edge of the tent more accurately.
At the same time, the farthest point is more accurate.
These results prove the effectiveness of the proposed BUBF and DCE.
The fourth scene is mainly occupied by a large wall.
Our method avoids wrongly predicting the door on the wall as the farthest region.
At the same time,
our method recovers richer shape details of small objects on the cabinet.
The fifth scene shows a lot of sundries.
Most methods fail to fully recover the shape details of the object.
Both our method and \cite{Chen2019SARPN} recover the details of small objects and the farthest area.
The last column shows a bathroom.
Although most methods give a more accurate farthest area, the edges of small objects generated by our method are sharper.
All the results above demonstrate that our method can achieve high accuracy on farthest region prediction and depth boundary recovery.

\subsection{Generalization Analysis on SUN-RGBD dataset}
\begin{figure}
	\centering
	\includegraphics[width=1\linewidth]{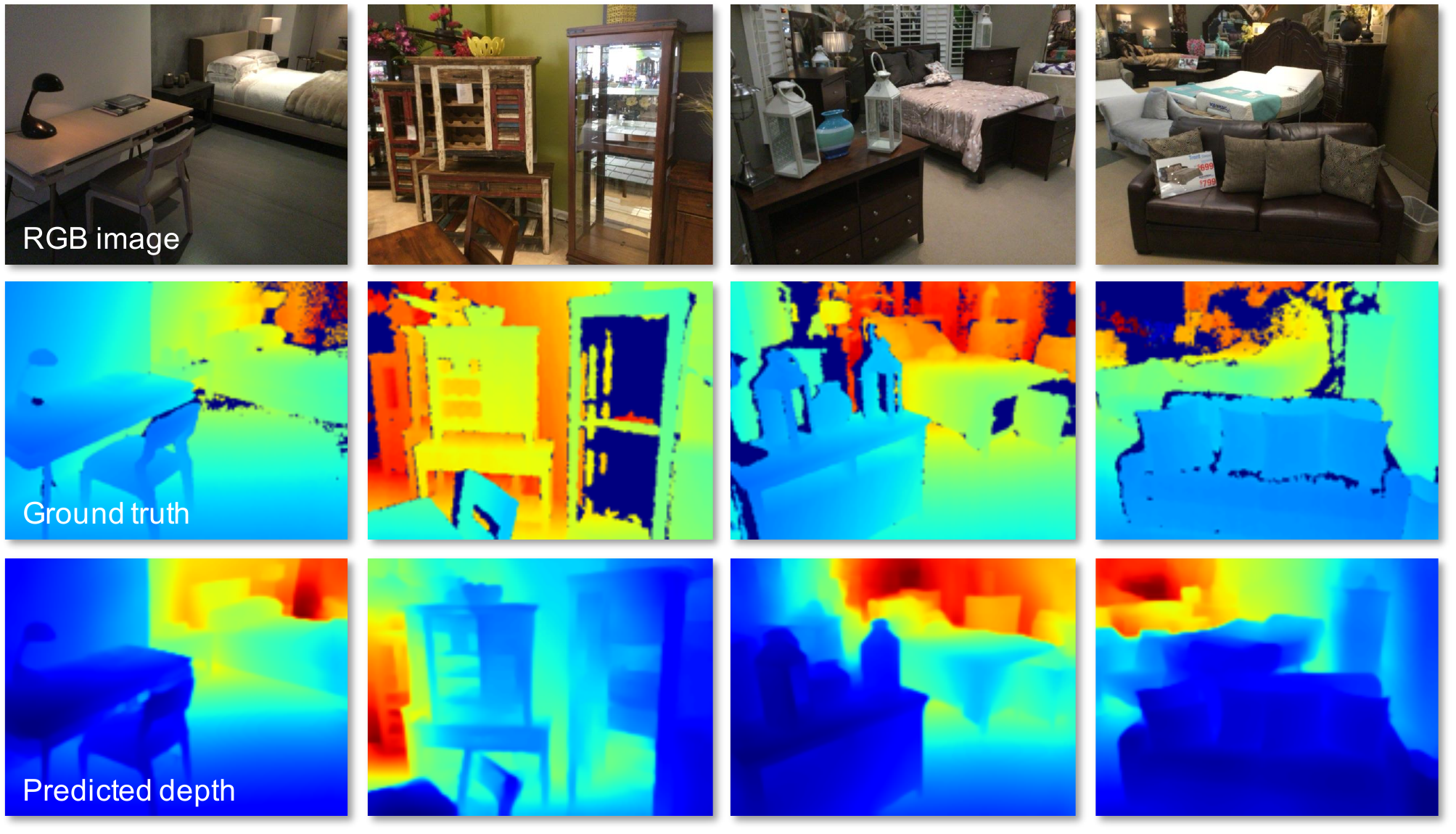}
	\caption{More visualization results on SUN-RGBD dataset.}
	\label{fig:gen}
\end{figure}

Besides the NYUD v2 dataset and the iBims-1 dataset,
another indoor dataset is employed to evaluate the generalization of our method.
Fig.\ref{fig:gen} demonstrates the visualization results on SUN-RGBD dataset \cite{sun},
and the predicted depth is generated by our network (trained on the NYUD v2 only).
Although the data distribution of the SUN-RGBD is totally different from that of NYUD v2,
our method achieves plausible results.
In detail,
regions that are relatively farther and closer are correctly predicted,
which means our method correctly predicts the structure in each scene.
Furthermore,
the predicted depth is reasonable where the depth sensor cannot capture the accurate value.

\subsection{Limitation and Future Work.}
\begin{figure}[!tb]
	\centering
	\includegraphics[width=0.7\linewidth]{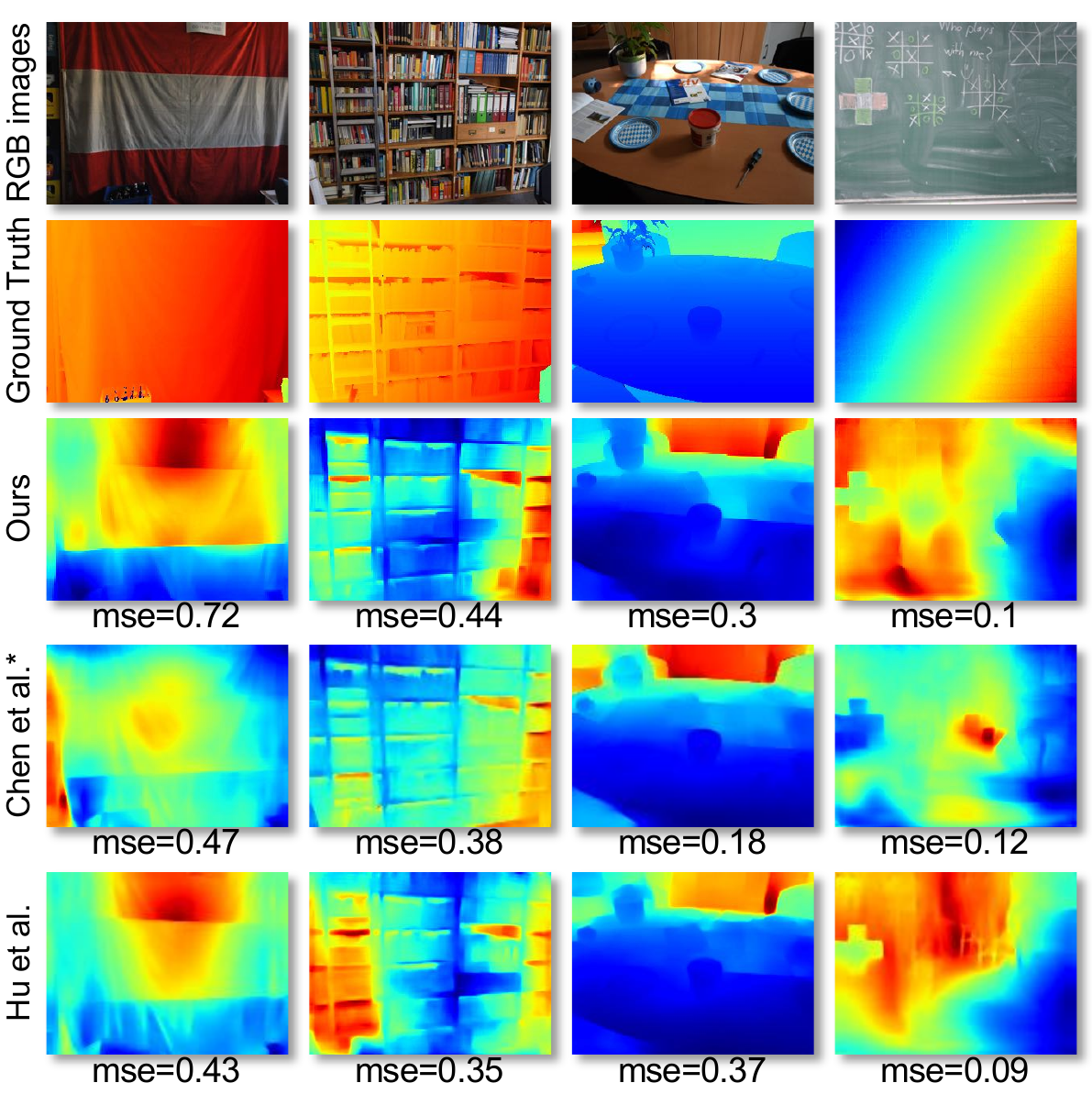}
	\caption{Several failure cases. Mse denotes the mean squared error between the predicted depth and the corresponding ground truth. * Using a ResNet-50 backbone, which is different from the original paper.}
	\label{fig:fail}
\end{figure}

In this paper,
the proposed method is to perceive the farthest region by capturing the contextual correlations and recover the depth boundary by fusing adjacent levels of the feature.
Thus, our approach handles the scenarios with rich visual cues well.
In contrast, the iBims-1 dataset contains several images that occupy a single large plane,
lacking the visual cue to assess the farthest area and depth.
Consequently,
in this case,
our method suffers from greater performance degradation than others,
limiting our method in the commonly used error metrics on the iBims-1 dataset.
This is the main limitation of our method.
Fig. \ref{fig:fail} shows four examples with a large plane occupying the whole image.
It can be seen that our method obtains a larger error than other methods in these cases.
In the future,
we are going to consider avoiding the incorrect depth change of the large plane by enhancing the feature embedding of the whole image.
Besides,
we also consider handling these cases by introducing additional semantic information into our network to constrain the depth change further.

\section{Conclusion}
In this paper,
we propose a novel Boundary-induced and Scene-aggregated network (BS-Net),
which considers the important roles of the farthest region and boundary cues in depth prediction.
To perceive the farthest region,
the DCE is introduced,
which captures the correlations between multi-scale regions.
To extract the important edge cue,
namely the boundary cue,
the BUBF is proposed to gradually locate sudden changes in depth without any other labels.
Besides,
the SRM is proposed to fully fuse the boundary cue and global depth layout. Numerous experiments on the NYUD v2 dataset indicate that our approach achieves state-of-the-art performance.

\section{Acknowledgment}
This work is supported by the National Natural Science Foundation of China No. 61703049.
This work is also supported by the BUPT Excellent Ph.D. Students Foundation, No. CX2020114,
the Natural Science Foundation of Hubei Province of China under Grant 2019CFA022.

\bibliography{Final_manuscript}
\end{document}